\documentclass{article}

\usepackage{arxiv}

\usepackage[utf8]{inputenc} 
\usepackage[T1]{fontenc}    
\usepackage{hyperref}       
\usepackage{url}            
\usepackage{booktabs}       
\usepackage{amsfonts}       
\usepackage{nicefrac}       
\usepackage{microtype}      
\usepackage{lipsum}		
\usepackage{graphicx}
\usepackage{natbib}
\usepackage{doi}
\usepackage{subcaption}
\usepackage{amsmath}
\usepackage{tabularray}
\usepackage{multirow}
\usepackage{algorithm}
\usepackage{algorithmicx}
\usepackage{algpseudocode}
\usepackage{amssymb}
\usepackage{appendix}
\usepackage{mathtools}
\usepackage{color}
 
\floatname{algorithm}{Algorithm}

\setcitestyle{numbers,square}

\title{Feature Norm Regularized Federated Learning: Transforming Skewed Distributions into Global Insights}

\author{ 
	\textbf{Ke HU} \\
	School of Cyber Science and Engineering\\
	Shanghai Jiao Tong University\\
	800 Dongchuan RD. Minhang District, Shanghai \\
	\texttt{crestiny@sjtu.edu.cn} \\
	\And
	\textbf{WeiDong QIU} \\
	School of Cyber Science and Engineering\\
	Shanghai Jiao Tong University\\
	800 Dongchuan RD. Minhang District, Shanghai \\
	\texttt{qiuwd@sjtu.edu.cn} \\
	\And
	\textbf{Peng TANG} \\
	School of Cyber Science and Engineering\\
	Shanghai Jiao Tong University\\
	800 Dongchuan RD. Minhang District, Shanghai \\
	\texttt{xxx@sjtu.edu.cn} \\
}

\renewcommand{\shorttitle}{\textit{arXiv} Template}

\hypersetup{
pdftitle={Feature Norm Regularized Federated Learning: Transforming Skewed Distributions into Global Insights},
pdfauthor={Ke HU, WeiDong QIU, Peng TANG},
pdfkeywords={Federated Learning, Feature Norm, Non-i.i.d Data, Data Heterogeneity},
}

\begin{document}
\maketitle

\begin{abstract}
	In the field of federated learning, addressing non-independent and identically distributed (non-i.i.d.) data remains a quintessential challenge for improving global model performance. This work introduces the Feature Norm Regularized Federated Learning (FNR-FL) algorithm, which uniquely incorporates class average feature norms to enhance model accuracy and convergence in non-i.i.d. scenarios. Our comprehensive analysis reveals that FNR-FL not only accelerates convergence but also significantly surpasses other contemporary federated learning algorithms in test accuracy, particularly under feature distribution skew scenarios. The novel modular design of FNR-FL facilitates seamless integration with existing federated learning frameworks, reinforcing its adaptability and potential for widespread application. We substantiate our claims through rigorous empirical evaluations, demonstrating FNR-FL's exceptional performance across various skewed data distributions. Relative to FedAvg, FNR-FL exhibits a substantial 66.24\% improvement in accuracy and a significant 11.40\% reduction in training time, underscoring its enhanced effectiveness and efficiency.
\end{abstract}


\section{Introduction}

Federated learning, a cutting-edge machine learning paradigm, has found extensive applications across domains such as healthcare, finance, and smart devices, enabling collaborative model training while preserving data privacy and security in the data-driven society \cite{banabilah2022federated}.
In federated learning, a crucial distinction from conducting deep learning on a single node is to aggregate locally updated models from individual participants to obtain a global model \cite{wang2019adaptive}. Most commonly used aggregation algorithms in federated learning is FedAvg \cite{mcmahan2017communication}. FedAvg performs weighted averaging on locally trained models to construct a global model, which is simple but notably effective under i.i.d. (independently and identically distributed) data \cite{clauset2011brief}.

Despite the widely application of FedAvg, a challenging case which arises in this domain is the non-i.i.d (non-identically and independently distributed) distributed  data \cite{hsieh2020non}. The challenging case leads to variations in local data characteristics among participants, hindering the convergence and generalization of global models.
Recent works \cite{li2019convergence} have mentioned that FedAvg has failed to perform well under non-i.i.d distributed data, because the weighted averaging cannot integrate proper and accurate knowledge from participants with skewed data distribution. 

Numerous efforts have since continued to improve the accuracy of global model in federated learning with non-i.i.d data distribution, such as FedProx\cite{li2020federated}, Scaffold\cite{karimireddy2020scaffold}, MOON\cite{li2021model} and FedNova\cite{wang2020tackling}. 
However, experiment results in \cite{li2022federated} have shown that these proposed algorithms does not greatly outperform from FedAvg under some challenging datasets such as CIFAR-10. Among image datasets, the classification task on CIFAR-10 is more difficult than on the others \cite{li2022federated}, thus resulting in poor accuracy of the global model. 

In this work we propose FNR-FL (Feature Norm Regularized Federated Learning), a federated learning framework leveraging the class average feature norm to enhance the performance of the global model under the non-i.i.d data distribution. Compared to prior efforts dedicated to addressing non-i.i.d. issues, FNR-FL allows for significantly better test accuracy and faster convergence and excels in various non-i.i.d. scenarios, instead of excelling in only a limited number of settings.

The main achievements, including contributions to the field can summarised as follows:
\begin{itemize}
	\item We proposed a regularization algorithm based on class average feature norms, which serves as a plug-and-play module that seamlessly integrates with existing federated learning algorithms. This modular algorithm allows for enhanced model regularization without the need for substantial modifications to the underlying federated learning framework, making it a versatile and easily adoptable solution.
	\item Building upon this regularization algorithm, we have constructed a federated learning algorithm FNR-FL that achieves significantly superior test accuracy and faster convergence under the non-i.i.d. data distribution. The FNR-FL algorithm demonstrated exceptional performance, particularly in feature distribution skew scenarios, achieving an accuracy of 0.9976 for ResNet-18, markedly superior to other FL (Federated Learning) algorithms.
	\item Under mixed non-i.i.d scenarios, FNR-FL has proven to outperform other algorithms, marking the first known federated learning algorithm tested under such mixed conditions. This strategic testing highlights the algorithm's robust performance in complex, real-world conditions that had not been considered by prior methodologies.
	\item We have developed two innovative metrics, $\kappa$ and $\rho$, to assess the efficiency and performance of FL algorithms. These metrics innovatively capture the trade-off between accuracy, communication, and computational costs, offering a clear benchmark for algorithm comparison. Our FNR-FL algorithm demonstrates its superiority through these metrics, ensuring high accuracy while maintaining low resource usage.
\end{itemize}

\section{Related work}

\subsection{Strategies for federated learning under non-i.i.d. data distribution}

Li et al. \cite{li2020federated} introduced FedProx, a novel optimization framework tailored for federated networks, uniquely addressing both system and statistical heterogeneity. Its innovation lies in allowing flexible local computations while incorporating a proximal term to maintain overall algorithmic stability across diverse devices. FedProx enables varying degrees of local computation on devices and employs a proximal term to enhance method stability.

Karimireddy et al. \cite{karimireddy2020scaffold} developed SCAFFOLD, a pioneering stochastic algorithm that creatively tackles gradient dissimilarity in federated settings. Its key innovation is the incorporation of control variates to significantly reduce the gradient variance, leading to enhanced convergence rates and model performance. The effectiveness of SCAFFOLD was confirmed by establishing robust convergence guarantees and conducting empirical evaluations.

Li et al. \cite{li2021model} proposed MOON (Model-Contrastive Learning), a groundbreaking approach in federated learning that introduces contrastive learning at the model level. This technique stands out for its simplicity and effectiveness, particularly in enhancing federated deep learning models' performance on non-IID datasets by encouraging model-level feature alignment.

Wang et al. \cite{wang2020tackling} introduced FedNova, a comprehensive theoretical framework designed for heterogeneous federated learning environments. Its main contribution is the natural integration of diverse local update steps and optimization techniques (such as GD, SGD, and proximal updates), ensuring fair and efficient convergence across a wide range of network conditions and device capabilities.

The FedDF framework \cite{lin2020ensemble} leverages a distillation technique to foster a more homogeneous knowledge transfer among decentralized datasets. This method is notable for its distillation from decentralized to centralized, aiding in overcoming the statistical heterogeneity inherent in FL.

FedGen \cite{zhu2021data} by Dong et al. integrates a generative component into the federated learning process. This addition aims to synthesize pseudo-samples that represent the global data distribution, thus enhancing the robustness of the model against data heterogeneity.

\subsection{Applications of feature norms}

Wei et al. \cite{wei2023edge} introduced cFedFN, a strategic approach designed to counteract the performance challenges posed by non-i.i.d. data in federated learning. This method innovatively clusters participants based on the similarity of their data distributions, effectively managing the diversity in data characteristics. The cornerstone of cFedFN lies in its unique use of feature norms: by employing a consistent feature extractor across all participants, cFedFN can accurately gauge the variations in data distributions. This is achieved by analyzing and comparing feature norms, which serve as reliable indicators of data distribution dissimilarities. This process not only assists in precise clustering but also significantly enhances the overall efficiency and accuracy of the federated learning model by aligning data characteristics more closely across different nodes.

\subsection{Cross-participant Knowledge Amalgamation}

Zhang et al. \cite{zhang2021parameterized} introduced KT-pFL, an advanced framework representing a significant leap in the domain of Personalized Federated Learning (pFL) through the integration of knowledge transfer. Distinct from traditional federated learning approaches, KT-pFL harnesses a knowledge coefficient matrix, a novel component that strategically amalgamates local soft predictions across different participants. Unlike conventional methods that often struggle with the diversity of participant data and learning models, KT-pFL addresses this challenge by enabling a tailored knowledge sharing mechanism. This approach marks a notable innovation in federated learning, offering a robust solution to the intricacies of personalized knowledge transfer and the challenges posed by participant data heterogeneity.

\section{Motivation}
\label{sec:others}

\subsection{The fundamental reason of non-i.i.d FL performance decline}

The primary cause of performance degradation in federated learning due to non-i.i.d data lies in the varied data distributions across different participants. This variation leads to divergent update directions when training local models \cite{zhao2018federated}. Simply put, the more diverse the data distribution, the greater the directional deviation in local model updates \cite{li2021fedbn}, which hampers the effective consolidation of knowledge into the global model \cite{karimireddy2020scaffold, li2022federated, wang2020tackling, gao2022feddc}. Consequently, this disparity in data distribution culminates in a decline in the global model's overall test accuracy.

\paragraph{Implication 1}
In light of the aforementioned perspective, our objective is to minimize discrepancies in model update directions among participants in federated learning, ultimately enhancing the overall performance of the global model.

\subsection{Quantify the difference among local model updates}

Acknowledging the critical role of minimizing update discrepancies in federated learning \textbf{(Implication 1)}, our focus shifts to quantifying these differences effectively. Wei et al. \cite{wei2023edge} suggest that by using a consistent feature extractor, it is feasible to approximate data distribution differences through the variance in feature norms.

\paragraph{Implication 2}
In a similar vein, we can postulate that, under the assumption of using the same dataset for evaluation, it is possible to gauge the difference among local model updates by leveraging the disparity among feature norms.

\subsection{Promoting Alignment of Model Update Directions Among Participants}

FedProx \cite{li2020federated}, introduced by Li et al., extends and redefines FedAvg to tackle the dual challenges of system and statistical heterogeneity, the latter pertaining to non-i.i.d data distributions among participants. To counteract the effects of non-i.i.d data, FedProx integrates a proximal term in each local model's loss function. This term penalizes deviations from the global model, aligning local updates more closely with it.

\paragraph{Implication 3}
Informed by FedProx's methodology, we propose enhancing participant training by integrating a regularization term based on feature norm differences \cite{brownlee2018gentle}. This term serves as a corrective measure for participants with diverse data distributions, encouraging more aligned contributions to the global model. 

\section{Non-i.i.d scenarios}

In our study, we have investigated three distinct categories of non-i.i.d settings: feature distribution skew, label distribution skew, and quantity skew \cite{zhang2022federated}. Additionally, we have explored two mixed non-i.i.d settings, which combine label distribution with quantity skew and label distribution with feature distribution skew. Subsequently, we will provide detailed descriptions of these three single non-i.i.d settings.

\subsection{Feature distribution skew}

\paragraph{Definition}

Feature distribution skew occurs when the marginal distributions of a feature $X$, as seen in $P_{train,1}(x), \cdots, P_{train,n}(x)$, vary across the training datasets of different participants $P_1,\cdots,P_n$. Conversely, the conditional distributions $P_{train,1}(y|x), \cdots, P_{train,n}(y|x)$ show little variation \cite{gao2023fedios}. This indicates that the skew is primarily in the input features $X$, while the output $Y$ maintains consistent conditional distributions. 

\paragraph{Simulation}

To emulate feature distribution skew, we introduce Gaussian noise to the data \cite{mou2023pfedv}. Gaussian noise is a continuous random variable whose probability distribution follows the Gaussian distribution (normal distribution), typically represented by the mathematical formula below:
\begin{equation}
p(x) = \frac{1}{\sqrt{2\pi}\sigma} e^{-\frac{(x - \mu)^2}{2\sigma^2}}
\label{Gaussian}
\end{equation}
where:
\begin{itemize}
	\item $x$ denotes the noise value.
	\item $\mu$ represents the mean.
	\item $\sigma$ signifies the standard deviation. A larger standard deviation results in a greater noise amplitude. 
\end{itemize}

The mathematical representation of generating data with feature distribution skew is as follows (where $\odot$ denotes element-wise multiplication):
\begin{equation}
x^{\prime} = x + \left( \varepsilon \odot M(x) \right) \cdot \sigma + \mu
\label{noise}
\end{equation}
where the mask $M(x)$ is a tensor of the same size as the input data $x$, with only specific elements set to 1 while the rest remain at 0. $\sigma$ represents the standard deviation of the Gaussian noise, and $\mu$ denotes the mean of the noise. The noise $\varepsilon$ is generated through Gaussian distribution $\mathcal{N} (0,1)$ and only affects the value masked by $M(x)$.

\paragraph{An example of feature distribution skew}

\begin{figure}
  \centering

  \begin{subfigure}{0.3\textwidth}
    \includegraphics[width=\textwidth]{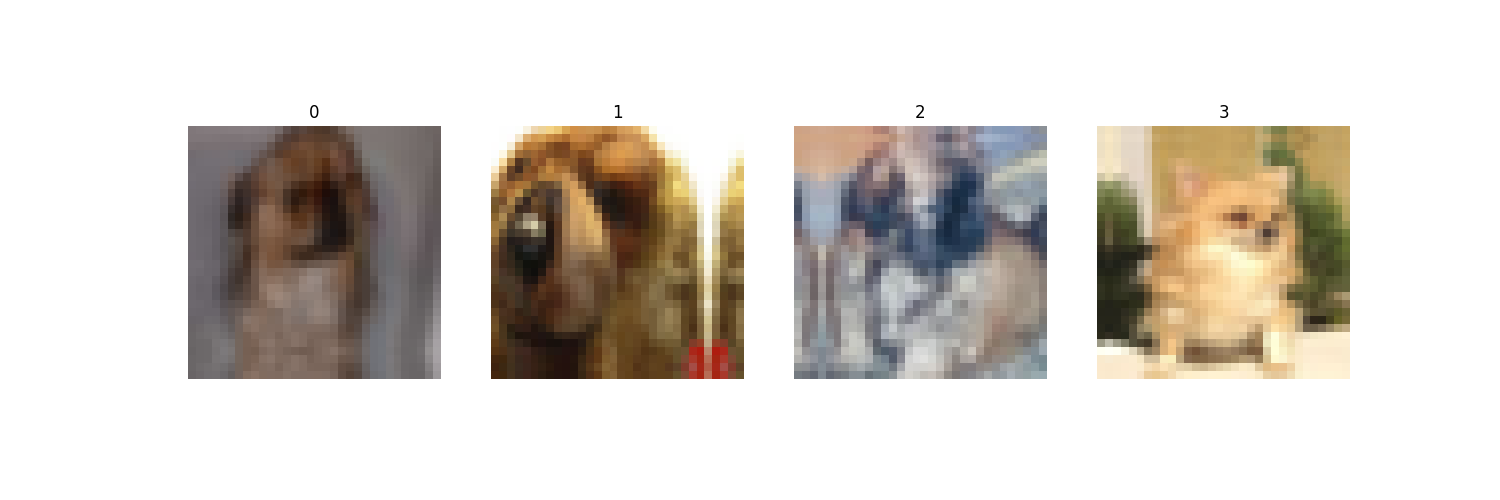}
    \caption{original samples}
	\label{origin}
  \end{subfigure}
  \begin{subfigure}{0.3\textwidth}
    \includegraphics[width=\textwidth]{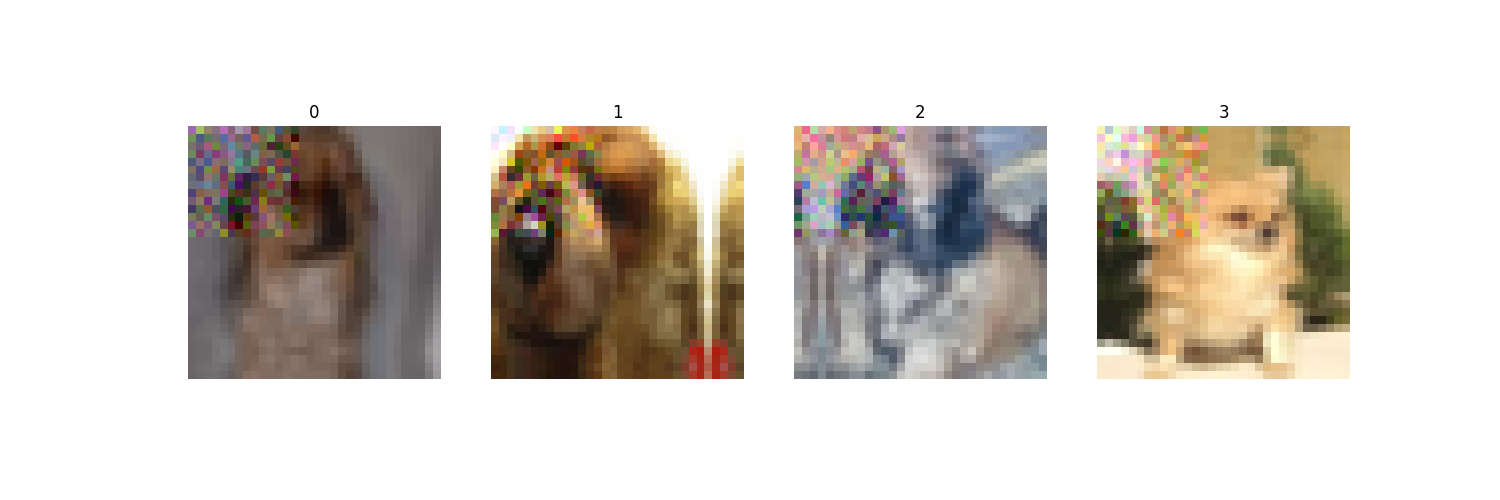}
    \caption{$noise \sim  Gau(0.1)$}
	\label{0.1}
  \end{subfigure}
  \begin{subfigure}{0.3\textwidth}
    \includegraphics[width=\textwidth]{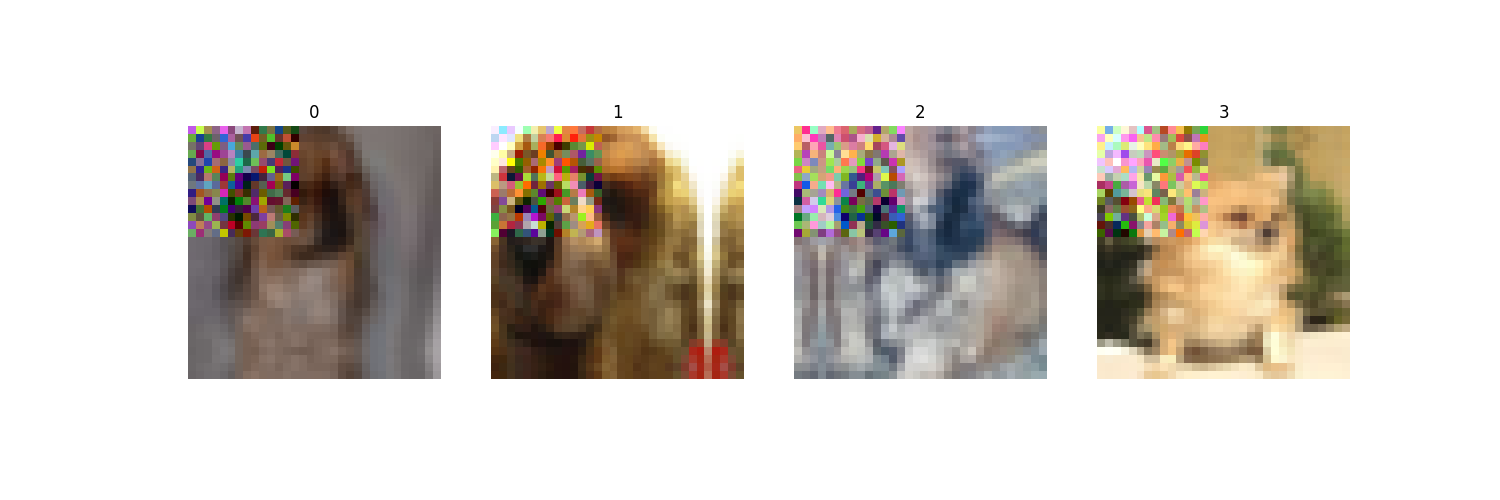}
    \caption{$noise \sim  Gau(0.2)$}
	\label{0.2}
  \end{subfigure}
  \begin{subfigure}{0.3\textwidth}
    \includegraphics[width=\textwidth]{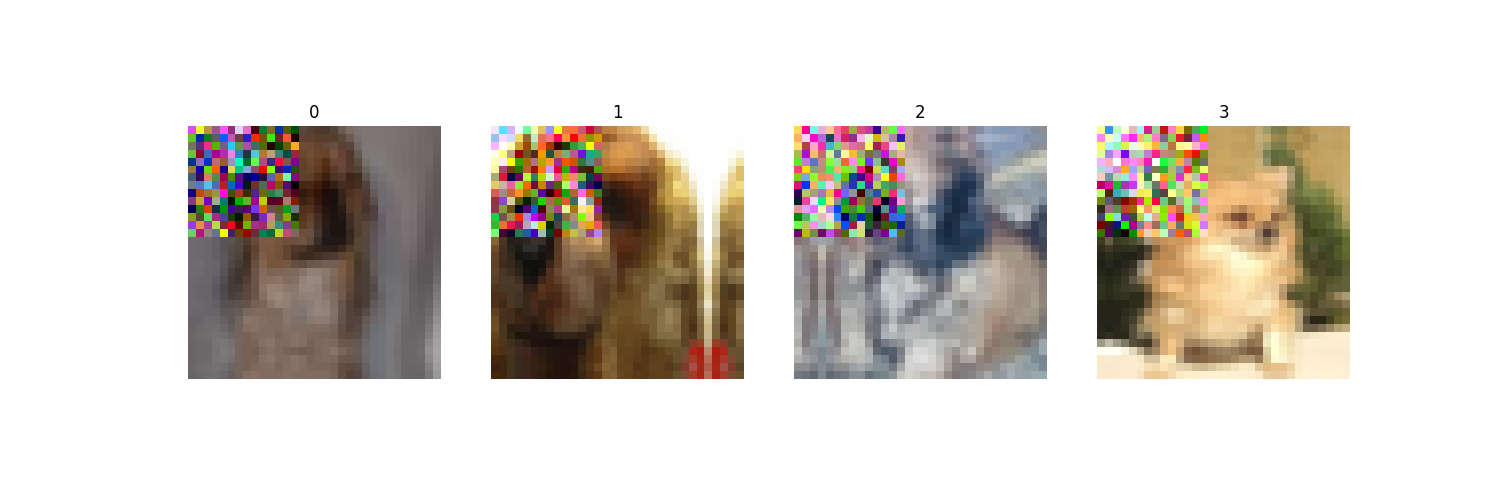}
    \caption{$noise \sim  Gau(0.3)$}
	\label{0.3}
  \end{subfigure}
  \begin{subfigure}{0.3\textwidth}
    \includegraphics[width=\textwidth]{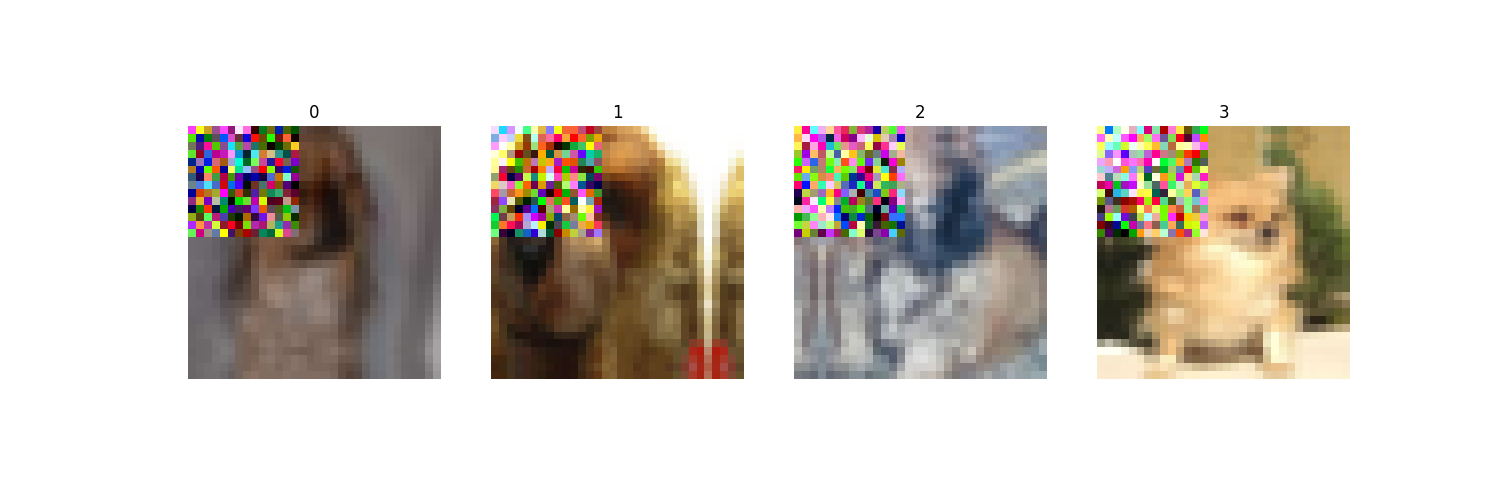}
    \caption{$noise \sim  Gau(0.4)$}
	\label{0.4}
  \end{subfigure}
  \begin{subfigure}{0.3\textwidth}
    \includegraphics[width=\textwidth]{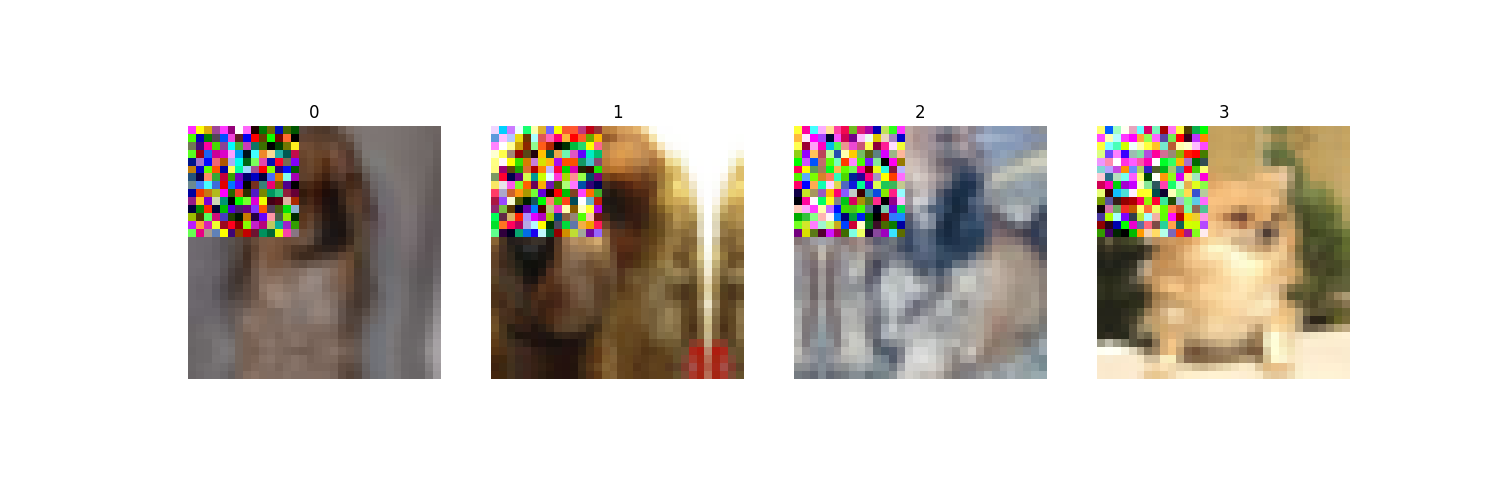}
    \caption{$noise \sim  Gau(0.5)$}
	\label{0.5}
  \end{subfigure}

  \caption{Visualization of feature distribution skew on CIFAR-10}

  \label{fig:FDS}
\end{figure}

We simulated feature distribution skew by adding Gaussian noise to the data held by each participant. Using CIFAR-10 dataset samples, we show in Figure \ref{fig:FDS} how applying Gaussian noise with varying standard deviations (0.1 to 0.5) affects the data. Original samples are in Figure \ref{origin}, while Figures \ref{0.1} to \ref{0.5} display the impact of increasing noise levels. As the standard deviation increases, the images become progressively obscured, exemplifying the feature distribution's deviation from its original form. 

\subsection{Label distribution skew}

\paragraph{Definition}

Label distribution skew \cite{zhang2022federated} is observed when the distribution of labels across various local datasets, denoted as $P_{train,1}(y), \cdots, P_{train,n}(y)$, differs among participants $P_1,\cdots,P_n$. This skewness occurs despite the similarity in conditional distributions $P_{train,1}(x|y), \cdots, P_{train,n}(x|y)$. Essentially, while the frequency of labels varies between datasets, the association between features and labels remains consistent. 

\paragraph{Simulation}

For label distribution skew, we employ the Dirichlet distribution to allocate data to different participants. The probability density function (PDF) of the Dirichlet distribution is defined as follows:

\begin{equation}
\text{Dir}(\mathbf{p} | \boldsymbol{\alpha}) = \frac{1}{\text{B}(\boldsymbol{\alpha})} \prod_{i=1}^{n} {(\frac{|D^{i}|}{\sum^{n}_{i=1}|D^{i}|})}^{\alpha_i - 1}
\label{dir}
\end{equation}
where:
\begin{itemize}
	\item $\mathbf{p} = [p_1, p_2, \ldots, p_{n}]$ represents a random vector, with each ${(\frac{|D^{i}|}{\sum^{n}_{i=1}|D^{i}|})}$ denoting the proportion of data allocated to the $i$-th participants.
	\item $\boldsymbol{\alpha} = [\alpha_1, \alpha_2, \ldots, \alpha_{n}]$ represents the parameter vector of the Dirichlet distribution. A higher $\alpha$ value leads to more skewed data distributions among participants.
	\item $\text{B}(\boldsymbol{\alpha})$ represents the multivariate beta function, used for normalization, defined as:
	\begin{equation}
	\text{B}(\boldsymbol{\alpha}) = \frac{\prod_{i=1}^{n} \Gamma(\alpha_i)}{\Gamma(\sum_{i=1}^{n} \alpha_i)}
	\label{beta}
	\end{equation}
	The choice of parameter vector $\boldsymbol{\alpha}$ can influence the uniformity of data allocation.
	\item Gamma function $\Gamma(\alpha) = \int_0^\infty t^{\alpha - 1} e^{-t} dt$ ensures that the probability density function integrates to 1. Proper scaling is vital for accurately representing the allocation of data across different participants.
\end{itemize}

In this experiment, the Dirichlet distribution parameters for all participants were set to the same value, denoted as $\alpha$. Consequently, the allocation of data to $n$ participants using a Dirichlet distribution with parameter $\alpha$ can be expressed as: $\mathbf{p} \sim Dir_n(\alpha)$. Sampling from the Dirichlet distribution yields a set of probability values $\mathbf{p}$ that represent the proportion of data allocated to different participants.

\paragraph{An example of label distribution skew}

\begin{figure}
	\centering

	\begin{subfigure}{0.4\textwidth}
		\includegraphics[width=\textwidth]{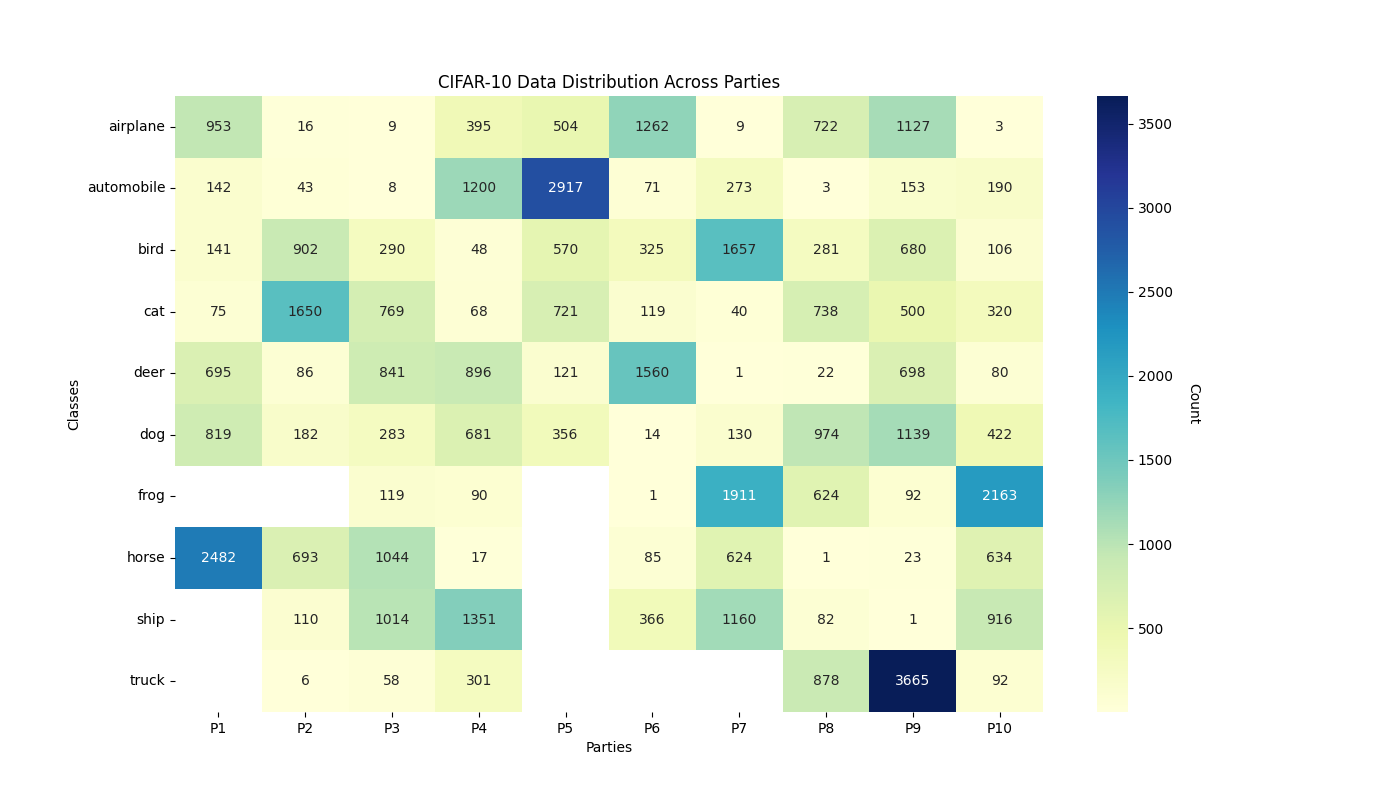}
		\caption{Label distribution skew}
		\label{LDS}
	\end{subfigure}
	\begin{subfigure}{0.4\textwidth}
		\includegraphics[width=\textwidth]{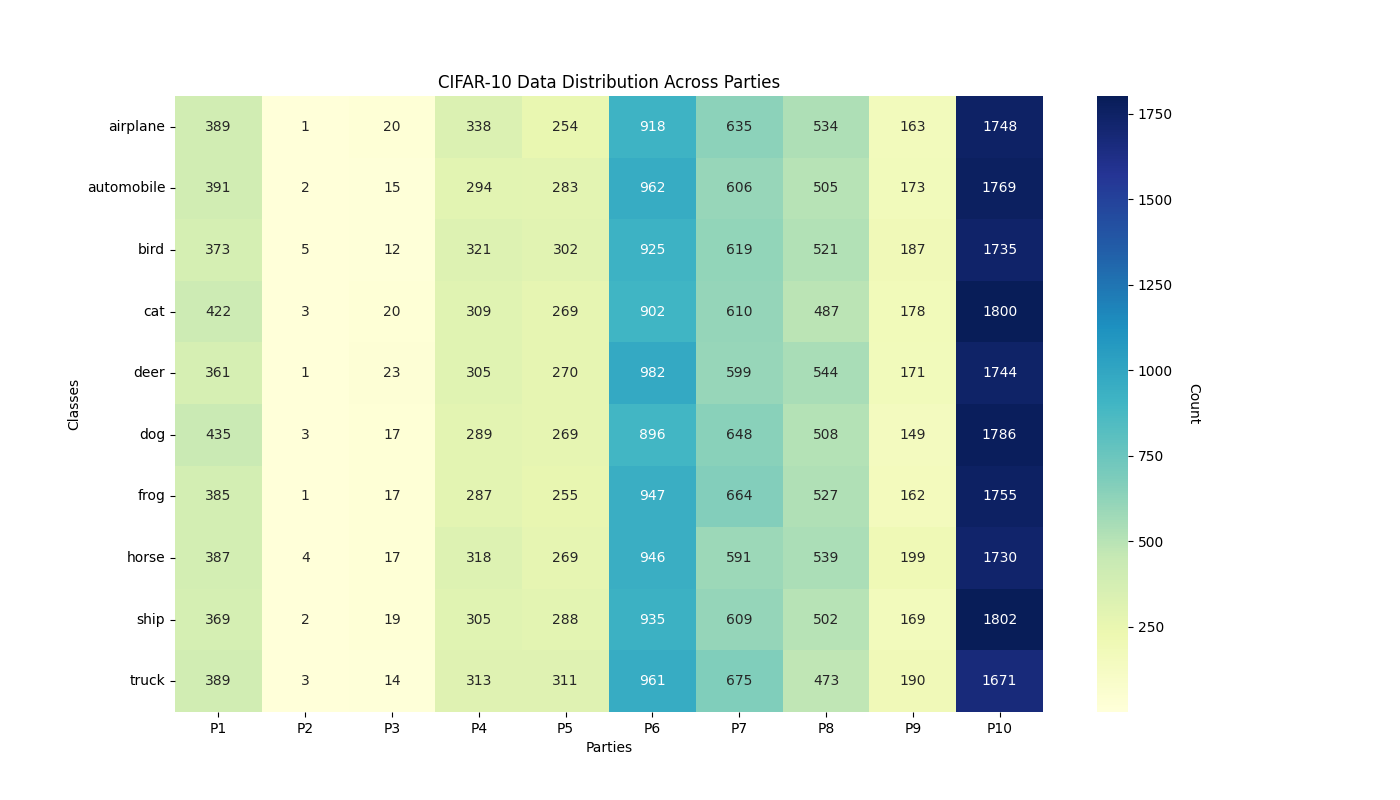}
		\caption{Quantity skew}
		\label{QS}
	\end{subfigure}
	\caption{Visualization of Label distribution skew and Quantity skew on CIFAR-10}

	\label{LDSQS}
\end{figure}

The Dirichlet distribution is utilized for allocating data with label distribution skew to different participants. Each class of samples is subject to a Dirichlet distribution sampling process individually. Taking the allocation of CIFAR-10 dataset to 10 participants as an example, when the parameter $\alpha$ is set to 0.5, the data allocation is depicted in the Figure \ref{LDS}. Darker colors indicate a higher quantity of samples, while white signifies the absence of samples in that class. From the illustration, it is evident that there are substantial variations in the number of samples from each class within the local datasets held by the participants. For instance, consider participant $P_5$, which possesses 2917 samples labeled as "automobile"; this represents an over-represented class. In contrast, there are no samples from classes such as "frog", "horse", "ship" and "truck" in $P_5$, indicating under-representation.

\subsection{Quantity skew}

Similar to label distribution skew, quantity skew arises when the sizes of local datasets, symbolized as $|D_i|$, differ among parties. We employ the Dirichlet distribution to distribute varying data sample quantities to each party. However, unlike label distribution skew, we maintain a roughly consistent data distribution across parties in this scenario \cite{archetti2023heterogeneous}. This consistency allows us to isolate and analyze the specific impact of quantity skew on global model performance. Figure \ref{QS} demonstrates this, showing that within each participant's local datasets, the counts of different labels fall within a specific range.

\section{Proposed Method}

\begin{table}[htbp]
	\centering
	\caption{Key Notations}
	  \begin{tabular}{cccccccc}
	  \toprule
	  & Notation & Semantics \\
	  \midrule 
	  & $n$ & number of participants \\
	  & $D_i$ & local dataset of participant $P_i$ \\
	  & $(x_{i}^{j}, y_{i}^{j})$ & $j$-th pair of sample and label of participant $P_i$  \\
	  & $M_i$ & local model of participant $P_i$ \\
	  & $Mf_i$ & feature extractor in local model of participant $P_i$ \\
	  & $U_{i}^{j}$ & feature norm $j$-th sample of participant $P_i$ \\
	  & ${\bf F}_{i}^{k}$ & class average feature norm of $k$-th class of participant $P_i$ \\
	  & $\Phi_{i}(w_{i}^{t})$ & objective function of participant $P_i$ in $t$-th round \\
	  & $L$ & the parameter of $L$-smooth \\
	  & $\mu$ & the parameter of $\mu$-strongly convex \\
	  & $\eta_m$ & learning rate in $m$-th mini-batch SGD \\
	  & $\bar{\mathbf{w}}_{m}$ & parameter of global model in $m$-th mini-batch SGD \\
	  & $\mathbf{w}^i_m$ & parameter of local model of participant $P_i$ in $m$-th mini-batch SGD \\
	  & $\mathbf{w}^*$ & parameter of optimal global model \\
	  & $b^i_m$ & a random batch selected from all batches of participant $P_i$ \\
	  \bottomrule
	  \end{tabular}%
	\label{notations}%
\end{table}%

\subsection{Class Average Feature Norm}

In this section, we describe the computation of class average feature norm. To facilitate our discussion, we enumerate key notations in Table \ref{notations}.
In a federated learning scenario for image classification tasks involving $n$ participants, each participant $P_i$ possesses a local dataset denoted as $D_i$ with $n_i$ labeled samples:
\begin{equation}
	D_{i} = \Big\{(x_{i}^{j}, y_{i}^{j})\Big\}_{j=1}^{n_{i}}
\end{equation}
 Datasets $D_i$ contains samples from $k_i$ different classes. Each participant's local model $M_i$ consists of two main components: the feature extractor $Mf_i$ and the classifier $Mc_i$. For a given sample $x^i_j$ in $D_i$, we define its feature norm as the Euclidean norm ($L_2$ norm) of the extracted features as \cite{naik2023exponential}:
\begin{equation}
	U_{i}^{j} = ||Mf_i(x_{i}^{j}, \vartheta_{f}^{i})||_2
\label{FN}
\end{equation}

where $\vartheta_{f}^i$ represents the parameters of the feature extractor $Mf_i$. After calculating the feature norms for each sample, we proceed to compute the average feature norm for each class of participant $P_i$. Specifically, for the class $k$ in participant $P_i$, the computation of class average feature norm ${\bf F}_{i}^{k}$ is performed as follows:
\begin{equation}
	{\bf F}_{i}^{k} = \frac{1}{\sum_{\mathit{j}=1}^{n_{i}} \delta_{[y_{i}^{j}=k]}}\sum_{\mathit{j}=1}^{n_{i}}\delta_{[y_{i}^{j}=k]}U_{i}^{j}
\label{CAFN}
\end{equation}
In this equation, $\delta _{[y_{i}^{j}=k]}$ is the indicator function, which is defined as:
\begin{equation}
	\delta _{[y_{i}^{j}=k]} = \begin{cases}
		1 & \text{if }\ y_i^j = k, \\
		0 & \text{if }\ y_i^j \neq k.
		\end{cases}
\end{equation}

This process allows us to obtain class average feature norms for the different classes of participants $P_i$. This computation enables us to extract meaningful insights about the feature representation across different classes within each participant's dataset. The computation of the class average feature norm is summarized in Algorithm \ref{CAFN_code}.

\begin{algorithm}
    \caption{Calculate Class Average Feature Norms}
    \begin{algorithmic}[1]
		\State \textbf{Input:} dataset $D$
        \State \textbf{Initialize:} class feature norms $F \gets $ \{\}, class sample counts $C$ $\gets$ \{\}, class average feature norms $avgF$ $\gets$ \{\}
        
        \For {each batch $B$ $in$ dataset $D$}
            
            \For {each sample ($x_i$, $y_i$) in batch $B$}
                \State compute the feature norm $U_i$ for $x_i$ using Eq. \eqref{FN}
                \State append $U_i$ to $F[y_i]$ 
                \State $C[y_i] += 1$
            \EndFor
        \EndFor
        
        \For {each (label $k$, feature norms $F_k$) $in$ $F$}
            \If {C[$k$] $>$ 0}
                \State Calculate the class average feature norm $avgf$ using Eq. \eqref{CAFN}
                \State $avgF[$k$]$ $\gets$ $avgf$
            \EndIf
        \EndFor
        
        \State \textbf{Output:} class average feature norms $avgF$
    \end{algorithmic}
	\label{CAFN_code}
\end{algorithm}

\subsection{FNR-FL: Feature Norm Regularized Federated Learning}

\subsubsection{Procedure of FNR-FL}

In this subsection, we introduce the FNR-FL algorithm, a novel federated learning approach designed to tackle the challenges posed by non-identically distributed (non-i.i.d) data across different participants. This framework leverages feature norm regularization to align local model updates, thus enhancing global model performance. The process involves each participant training their local model, computing class average feature norms, and then adjusting their models based on these norms to achieve better alignment with the global model. The detailed steps of this procedure are outlined in Algorithm \ref{FNR-FL}.

\paragraph{Local Training with computation of class average feature norm}

In each communication round $t \in [T]$,  all $n$ participants are active. The server sends the global model $w^t$ to each participant. Each participant $P_i$ then updates its local model $w_{i}^{t}$ on local dataset $D_i$ for $E_{train}$ epochs. Subsequently, each participant is required to compute the class average feature norms $F^t_i$ (Algorithm \ref{CAFN_code}) and the accuracy $A_i$ of samples in the public dataset $D_{public}$ using its updated local model $w_i^{t}$. The updated local model $w_i^t$, the class average feature norms $F_i^t$ and the accuracy $A_i$ are then sent to the server.

\paragraph{Computation of differences among feature norms}

Upon receiving the test accuracies, denoted as $A_{1},..., A_{n}$, on the public dataset $D_{public}$, the server proceeds to select $n*p$ participants with the lowest accuracies to undergo feature norm regularization. Here, $p$ serves as a threshold parameter determining the proportion of participants chosen for regularization. The set of these selected participants is denoted as $S_{re}$ and the set of all participants is denoted as $S$. For each participant $P_j$ in $S_{re}$, the server computes the differences of the class average feature norms between $P_j$ and other participants in $S \setminus S_{re}$ using Algorithm \ref{Diff}.

\paragraph{Feature Norm Regularized Model Update}

Consequently, the loss function $\Phi_{j}(w_{j}^{t})$ of participant $P_j$ incorporates its own cross-entropy loss $L(w^{k}_{i},B)$ and the difference of the class average feature norms between it and other participants $J(j, F_n)$:
\begin{equation}
	\Phi_{j}(w_{j}^{t})=\underbrace{L(w^{k}_{j},B)}_{Cross-Entropy\ Loss}+\underbrace{\lambda \cdot J(j, F_n)}_{Regularization\ Term }
\end{equation}
Here, the hyperparameter $\lambda$ balances the impact of local cross-entropy loss with the goal of reducing discrepancies in feature norms. In essence, $\lambda$ controls the trade-off between fitting the local datasets well and encouraging convergence towards a global model with better performance.

\begin{algorithm}
    \caption{Calculate Feature Norms Differences}
	\begin{algorithmic}[1]
		\State \textbf{Input:} participant $P_j$, set of all participants $S$, set of participants to be regularized $S_{re}$, feature norm list of all participants $F_n$
		\State $\Delta_{sum} = \{\}$
		\For {$m \in S \setminus S_{re}$ }
			\For {label $l$ $\in$ $F_n[P_j].keys()$}
				\State $\Delta = (F_n[m][l] - F_n[j][l])$
				\State $\Delta_{sum}[l] \mathrel{+}= \Delta$
			\EndFor
		\EndFor
		\State return $\Delta_{sum}$

	\end{algorithmic}
	\label{Diff}
\end{algorithm}

\begin{algorithm}
    \caption{FNR-FL}
    \begin{algorithmic}[1]
		\State \textbf{Input:} local datasets $D_i$, number of participants $n$, number of communication rounds $T$, number of local training epochs $E_{train}$, number of regularization epochs $E_{re}$, public dataset $D_{public}$, learning rate $\eta $, percentage of regularized participants $p$, 
        \State \textbf{Initialize:} global model $w^0$

		\State \textbf{Server execues:}
        \For {$t = 1, 2, \cdots, T$}
            
			initialize feature norm list $F_n$
            \For {$i = 1, 2, \cdots, n$ \textbf{in parallel}}
                \State send the global model $w^t$ to participant $P_i$ 
                \State $w_i^t, F_i^t, accuracy \gets \textbf{LocalTraining}(i, w^t, D_i)$
				\State $F_n[i] = F_i^t$ 
            \EndFor

			\State choose $n_{re} = \left\lfloor n*p \right\rfloor $ participants to refine local model updates

			\For {$j = 1, 2, \cdots, n_{re}$ \textbf{in parallel}}
				\State compute $\Delta_j$ using Algorithm \ref{Diff}
				\State $w_j^t \gets \textbf{FeatureNormRefine}(j, w_j^t, D_j, \Delta_j)$

			\EndFor
			\State $w^{t+1}\leftarrow \sum^{n}_{i=1}\frac{|D^{i}|}{\sum^{n}_{i=1}|D^{i}|}w_{i}^{t}$

        \EndFor
		\State return $w^T$
        \vspace{12pt}
		\State \textbf{LocalTraining}($i, w^t, D_i$):
		\State $w^t_i \gets w^t$
        \For {$k = 1, 2, \cdots, E_{train}$}
			\For {each batch $B$ $in$ dataset $D_i$}
				\State $L(w_{i}^{k};{B}) = \sum_{(x,y) \in B}^{}\ell(w_i^k;x;y)$
				\State $w_{i}^{k}\,\leftarrow\,w_{i}^{k}-\eta\nabla L(w_{i}^{k};{B})$
			\EndFor
        \EndFor
		
		\State compute $F_i^t$ on $D_{public}$ using Algorithm \ref{CAFN_code}
		\State test $w_{i}^{k}$ on $D_{public}$ and record accuracy $A_i$

		\State return $w_{i}^{k}, F_i^t, A_i$

		\vspace{12pt}
		\State \textbf{FeatureNormRegularization}($j, w_j^t, D_j, \Delta_j$):
        \For {$k = 1, 2, \cdots, E_{re}$}
			\For {each batch $B$ $in$ dataset $D_{public}$}
				\State $L(w_{j}^{k};{B}) = \sum_{(x,y) \in B}^{}\ell(w_j^k;x;y) $
				\State denote unique labels in $B$ as $L$
				\State $J(j, F_n) = 0.0$
				\For {each label $l \in L$}
					\State $\rho = \#(y==l)/|B|$
					\State $J(j, F_n) \mathrel{+}=\rho \cdot \Delta_j[l]$
				\EndFor
				\State $\Phi (w_j^k) = L(w_{j}^{k};{B}) + \lambda \cdot J(j, F_n)$
				\State $w_{j}^{k}\,\leftarrow\,w_{j}^{k}-\eta\nabla \Phi (w_j^k)$
			\EndFor
        \EndFor

		\State return $w_{j}^{k}$ to the server

    \end{algorithmic}
	\label{FNR-FL}
\end{algorithm}

\begin{figure}
	\centering
	\includegraphics[width=1.0\textwidth]{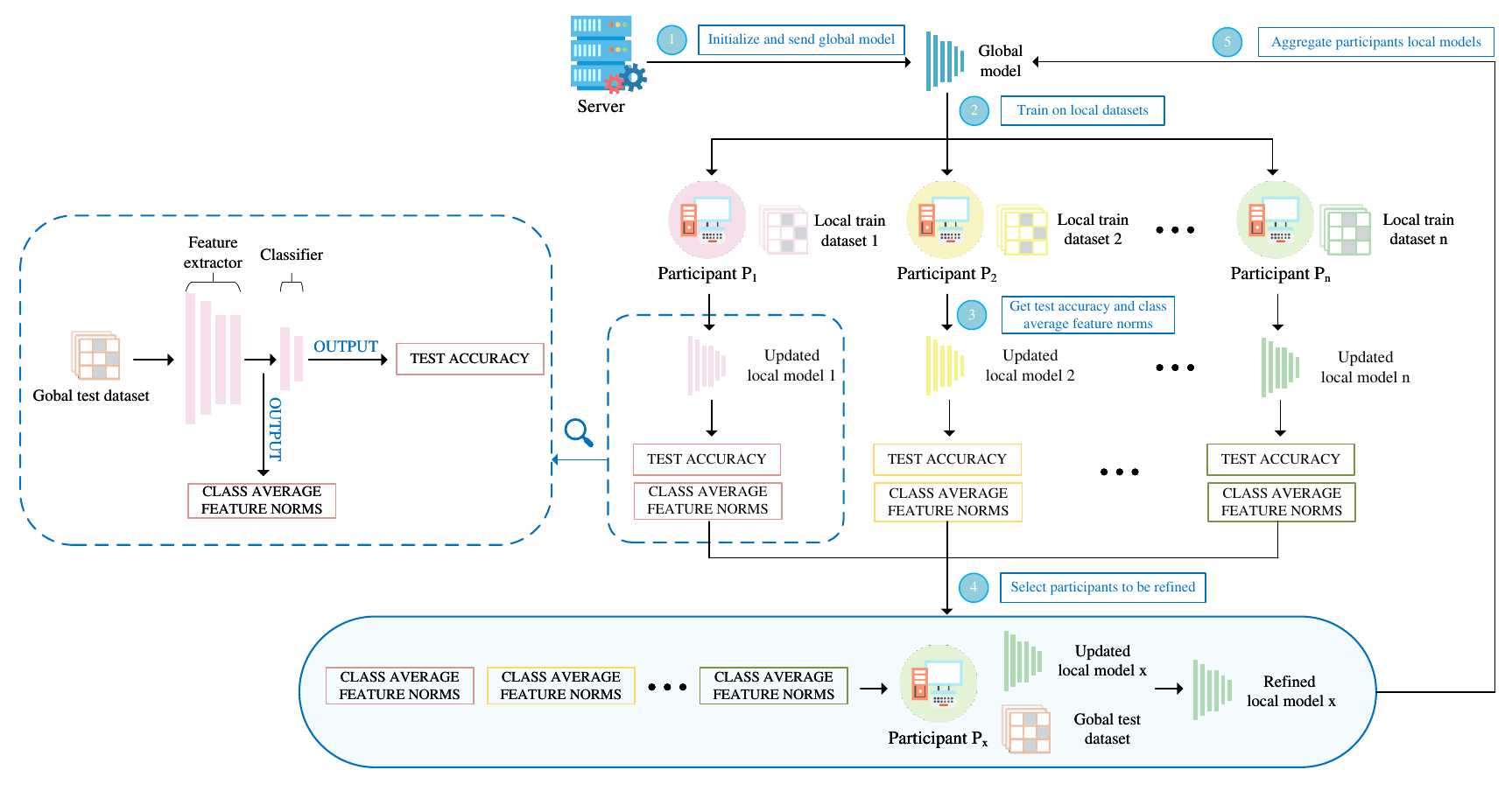} 
	\caption{The framework of FNR-FL.}
	\label{system}
\end{figure}

\subsubsection{Convergence Analysis of FNR-FL}

In this section, we delineate the convergence characteristics of the FNR-FL algorithm we proposed. Initially, we establish certain assumptions, commonly adopted in preceding studies \cite{zhou2018fenchel, bao2022fast, t2020personalized, haddadpour2019convergence}.

\begin{itemize}
    \item[\textbf{1.}] The functions $\Phi_1, \ldots, \Phi_N$ are all $L$-smooth. This implies that for all $x, y$, the following inequality holds:
    \begin{equation}
        \Phi_i(y) \leq \Phi_i(x) + \langle \nabla\Phi_i(x), y-x \rangle + \frac{L}{2} \| y-x \|^2.
    \end{equation}
    
    \item[\textbf{2.}] The functions are all $\mu$-strongly convex, which means for all $x, y$:
    \begin{equation}
        \Phi_i(y) \geq \Phi_i(x) + \langle \nabla\Phi_i(x), y-x \rangle + \frac{\mu}{2} \| y-x \|^2.
    \end{equation}
    
    \item[\textbf{3.}] The stochastic gradients are unbiased and have bounded variance. Specifically, in $m$-th mini-batch gradient descent step of participant $P_i$:
    \begin{equation}
        \mathbb{E} \left[ \| \nabla \Phi_i(w^i_{m}, b^i_{m}) - \nabla \Phi_i(w^i_{m}) \|^2 \right] \leq ({\Delta G}_i)^2,
    \end{equation}
    and
    \begin{equation}
        \mathbb{E} \left[ \| \nabla \Phi_i(w^i_{m}, b^i_{m}) \|^2 \right] \leq G^2.
    \end{equation}

	where $b^i_m$ denotes a random batch in all batches of participant $P_i$ in $m$-th mini-batch gradient descent step.
\end{itemize}

Consider the $m$-th mini-batch gradient descent step: $\bar{\mathbf{w}}_{m+1} = \bar{\mathbf{w}}_m - \eta_m \mathbf{g}_m$. We analyze the distance to the optimal point $\mathbf{w}^*$ in terms of the squared norm:
\begin{equation}
	\begin{aligned}
		\|\bar{\mathbf{w}}_{m+1} - \mathbf{w}^*\|^2 \\
		&= \|\bar{\mathbf{w}}_m  -\mathbf{w}^* - \eta_m \bar{\mathbf{g}}_m \|^2 + \eta_m^2 \|\bar{\mathbf{g}}_m - \mathbf{g}_m\|^2 
		+ 2\eta_m \langle \bar{\mathbf{w}}_m - \mathbf{w}^* - \eta_m \bar{\mathbf{g}}_m, \bar{\mathbf{g}}_m - \mathbf{g}_m \rangle.
	\end{aligned}
	\label{_i}
\end{equation}

where $\mathbf{g}_m = \nabla\Phi_i(\mathbf{w}^i_m, b^i_{m})$, which denotes the gradient of the loss function $\nabla\Phi_i$ with respect to the current mini-batch $b_m^i$; and its average $\bar{\mathbf{g}}_m = \nabla\Phi_i(\mathbf{w}^i_m)$.

\paragraph{\textit{Lemma 1}} 
The bound of $\|\bar{\mathbf{w}}_m - \mathbf{w}^* - \eta_m \bar{\mathbf{g}}_m\|^2$: 

\begin{equation}
	\begin{aligned}
		\MoveEqLeft \|\bar{\mathbf{w}}_m - \mathbf{w}^* - \eta_m \bar{\mathbf{g}}_m\|^2  \\
		&\leq (1-\mu\eta_m) \|\bar{\mathbf{w}}_m - \mathbf{w}^*\|^2 + 2 \sum_{i=1}^{n} {(\frac{|D^{i}|}{\sum^{n}_{i=1}|D^{i}|})} \|\bar{\mathbf{w}}_m - \mathbf{w}^i_m\|_2^2 + \frac{3\Gamma}{8L}.
	\end{aligned}
	\label{_g}
\end{equation}

where $\Gamma = \Phi(\mathbf{w}^*) - \sum_{i=1}^{n} {(\frac{|D^{i}|}{\sum^{n}_{i=1}|D^{i}|})} \Phi_i(\mathbf{w}^*)$.

Lemma 1 provides an upper bound of $\|\bar{\mathbf{w}}_m - \mathbf{w}^* - \eta_m \bar{\mathbf{g}}_m\|^2$ characterized by the learning rate $\eta_m$, $\mu$ and $L$.

\paragraph{\textit{Lemma 2}}
In order to associate with our assumptions, we consider the expectation of $\|\bar{\mathbf{g}}_m - \mathbf{g}_m\|^2$:

\begin{equation}
	\begin{aligned}
		\mathbb{E} \|\bar{\mathbf{g}}_m - \mathbf{g}_m\|^2 
		&\leq \sum_{i=1}^{n} {(\frac{|D^{i}|}{\sum^{n}_{i=1}|D^{i}|})}^2 ({\Delta G}_i)^2 \quad \text{(by Assumption 3)}.
	\label{l}
	\end{aligned}
\end{equation}

Lemma 2 bounds the expectation of the gradient difference, which is influenced by the variability in each mini-batch's gradient, as indicated by Assumption 3.

\paragraph{\textit{Lemma 3}}
Based on Lemma 1 and Lemma 2, we get:

\begin{equation}
	\begin{aligned}
		\mathbb{E} \|\bar{\mathbf{w}}_{m+1} - \mathbf{w}^*\|^2 
		&\leq (1-\mu\eta_m) \mathbb{E} \|\bar{\mathbf{w}}_m - \mathbf{w}^*\|^2 + 8 \eta_m^2 (E_{re} - 1)^2 G^2 + \eta_m^2 \sum_{i=1}^{n} {(\frac{|D^{i}|}{\sum^{n}_{i=1}|D^{i}|})}^2 ({\Delta G}_i)^2 + \frac{3\Gamma}{8L}.
	\end{aligned}
\end{equation}

Lemma 3 introduces a recursive relationship derived from the previous lemmas. This relation, expressed in terms of the expected squared norm difference between the current and optimal weight vectors, encapsulates the rate of convergence and gradient fluctuation, offering a predictive view of the model's refinement over iterations.

Letting $\delta_{m} = \mathbb{E} \|\bar{\mathbf{w}}_m - \mathbf{w}^*\|^2$, which denotes the expected squared norm of the deviation of the global model $\bar{\mathbf{w}}_m$ from the optimal ${\mathbf{w}}^*$ at the $m$-th mini-batch SGD; and thus $\delta_{m+1} = \mathbb{E} \|\bar{\mathbf{w}}_{m+1} - \mathbf{w}^*\|^2$, we can get the following recurrence relation:
\begin{equation}
	\begin{aligned}
		\delta_{m+1}
		&\leq (1-\mu\eta_m) \delta_{m} + 8 \eta_m^2 (E_{re} - 1)^2 G^2 + \eta_m^2 \sum_{i=1}^{n} {(\frac{|D^{i}|}{\sum^{n}_{i=1}|D^{i}|})}^2 ({\Delta G}_i)^2 + \frac{3\Gamma}{8L}.
	\end{aligned}
	\label{_q}
\end{equation}

$\delta_{m}$ stands for the expectation of the difference between global model in the $m$-th mini-batch SGD $\bar{\mathbf{w}}_m$ and the optimal global model $w^*$. 
This inequality is instrumental in demonstrating the decrement of $\delta_m$ over iterations, thus assuring convergence to the optimal global model $w^*$ as $m$ increases.

In our convergence analysis, we have deduced that for the sequence $\{\delta_m\}$, representing the expected deviation of the global model from the optimal during $m$-th mini-batch SGD, to converge, it is imperative that specific conditions regarding the learning rate and gradient norms are met. The requisite conditions are enumerated as follows:

\begin{enumerate}
	\item The learning rate $\eta_m$ decays with the number of mini-batch SGD.
	\item $0 < (1 - \mu\eta_m) < 1$
	\item $(E_{re} - 1)^2 G^2$, $({\Delta G}_i)^2$ and $\frac{3\Gamma}{8L}$ are bounded to ensure that the deviations $\delta_m$ do not grow unbounded which would jeopardize convergence.
\end{enumerate}

Given that all the prescribed conditions are satisfied, we ascertain that $\delta_m$ convergences to 0 as $m$ increases to infinity, signifying the assured convergence of the global model $\bar{\mathbf{w}}_m$ to the optimal global model $w^*$. 
While in practice the number of iterations \( m \) is finite, the convergence behavior of \( \delta_m \) as outlined still assures that we can approximate the desired precision within our computational budget.
For the complete proof, see the Appendix \ref{Proof}.

\section{Evaluations}

\subsection{Experimental Setup}

In our experimental setup, all participants are actively involved in every round of the training process. We use the SGD optimizer \cite{robbins1951stochastic} with learning rate 0.1. The batch size of the training data is set to 64 and the test data is set to 32 by default. The number of local training epochs is set to 10 by default and the regularization epochs is set to 5 by default. We conduct each training process for 10 rounds.

\subsection{Performance Under Single Types of Non-i.i.d Scenarios}

To investigate the effectiveness of the proposed FNR-FL, we report the performance of FNR-FL, FedAvg, FedProx, SCAFFOLD, MOON and FedNova when training ResNet-18 \cite{he2016deep} or VGG-11 \cite{krizhevsky2009learning} model on CIFAR-10 \cite{krizhevsky2009learning}. Given the formidable challenges posed by CIFAR-10 in non-i.i.d scenarios \cite{li2022federated}, we opted to employ the CIFAR-10 dataset in our experiments to assess the performance of FNR-FL. Experiments are conducted under three single types of non-i.i.d data distribution,  including feature distribution skew, label distribution skew and quantity skew.

\subsubsection{Test Accuracy}

The test accuracy results of training ResNet-18 or VGG-11 on CIFAR-10 under three single types of non-i.i.d scenarios are presented in Table \ref{Exp1}. We have the following analysis in aspect of different non-i.i.d scenarios:

\begin{enumerate}
	\item Outstanding in Feature Distribution Skew: Within the feature distribution skew scenario, the FNR-FL algorithm demonstrates remarkable effectiveness, significantly outperforming other FL methods. It achieves impressive accuracies of 0.9976 on ResNet-18 and 0.9931 on VGG-11, starkly surpassing other algorithms whose accuracies range between 0.5530 and 0.7425. This performance highlights FNR-FL's strong capability in handling datasets with feature distribution skew.
	\item Excellence in Label Distribution Skew: In label distribution skew, FNR-FL continues to excel, recording an accuracy of 0.9970 for ResNet-18 and an almost perfect score of 0.9992 for VGG-11. These results emphasize the algorithm's robustness in environments with skewed label distributions.
	\item Consistent Performance in Quantity Skew: Facing quantity skew, the FNR-FL algorithm still performs well on ResNet-18 and VGG-11. This illustrates the algorithm's consistent strength across diverse data distributions.
\end{enumerate}

In essence, the FNR-FL algorithm stands out as a highly effective solution in federated learning, especially in tackling the complexities of feature and label distribution skews. Its consistent high performance across various non-i.i.d scenarios marks its superiority  and robustness in the field of federated learning.
  
  \begin{table}[htbp]
	\centering
	\caption{Test accuracy on CIFAR-10 under single types of non-i.i.d scenarios}
	  \begin{tabular}{c|cc|cc|cc}
	  \toprule
	  \multicolumn{1}{c|}{\textbf{Non-i.i.d scenarios}} & \multicolumn{2}{c|}{Feature distribution skew} & \multicolumn{2}{c|}{Label distribution skew} & \multicolumn{2}{c}{Quantity skew} \\
	  \midrule
	  Model & ResNet-18 & VGG-11 & ResNet-18 & VGG-11 & ResNet-18 & VGG-11 \\
	  \midrule 
	  FNR-FL & \textbf{0.9976 } & \textbf{0.9931 } & \textbf{0.9970 } & \textbf{0.9992 } & \textbf{0.9982 } & \textbf{0.9939 } \\
	  FedAvg \color{blue}{(AISTATS, 2017 \cite{mcmahan2017communication})} & 0.6001  & 0.7485  & 0.8393  & 0.8311  & 0.9107  & 0.9194 \\
	  FedProx \color{blue}{(MLsys, 2020 \cite{li2020federated})} & 0.5956  & 0.7285  & 0.8773  & 0.8050  & 0.8874  & 0.8916 \\
	  SCAFFOLD \color{blue}{(ICML, 2020, \cite{karimireddy2020scaffold})} & 0.7425  & 0.7733  & 0.9077  & 0.8657  & 0.7017  & 0.9064 \\
	  MOON \color{blue}{(CVPR, 2021, \cite{li2021model})} & 0.5530  & 0.7193  & 0.6515  & 0.8432  & 0.7336  & 0.8424 \\
	  FedNova \color{blue}{(NeurIPS, 2020, \cite{wang2020tackling})} & 0.6223  & 0.7475  & 0.9043  & 0.8522  & 0.7293  & 0.7622 \\
	  FedDyn \color{blue}{(ICPADS, 2023, \cite{jin2023feddyn})} & 0.6071  & 0.7555  & 0.8463  & 0.8381  & 0.9177  & 0.9264 \\
	  FedFTG \color{blue}{(CVPR, 2022, \cite{zhang2022fine})} & 0.6451  & 0.7935  & 0.8843  & 0.8761  & 0.9557  & 0.9644 \\
	  FedDF \color{blue}{(NeurIPS, 2020, \cite{lin2020ensemble})} & 0.6101  & 0.7585  & 0.8493  & 0.8411  & 0.9207  & 0.9294 \\
	  FedDC \color{blue}{(CVPR, 2022, \cite{gao2022feddc})} & 0.6401  & 0.7885  & 0.8793  & 0.8711  & 0.9507  & 0.9594 \\
	  FedGen \color{blue}{(ICML, 2021, \cite{zhu2021data})} & 0.6021  & 0.7505  & 0.8413  & 0.8331  & 0.9127  & 0.9214 \\
	  \bottomrule
	  \end{tabular}%
	\label{Exp1}%
\end{table}%

\subsubsection{Convergence}

\begin{figure}
	\centering

	\begin{subfigure}{0.3\textwidth}
		\includegraphics[width=\textwidth]{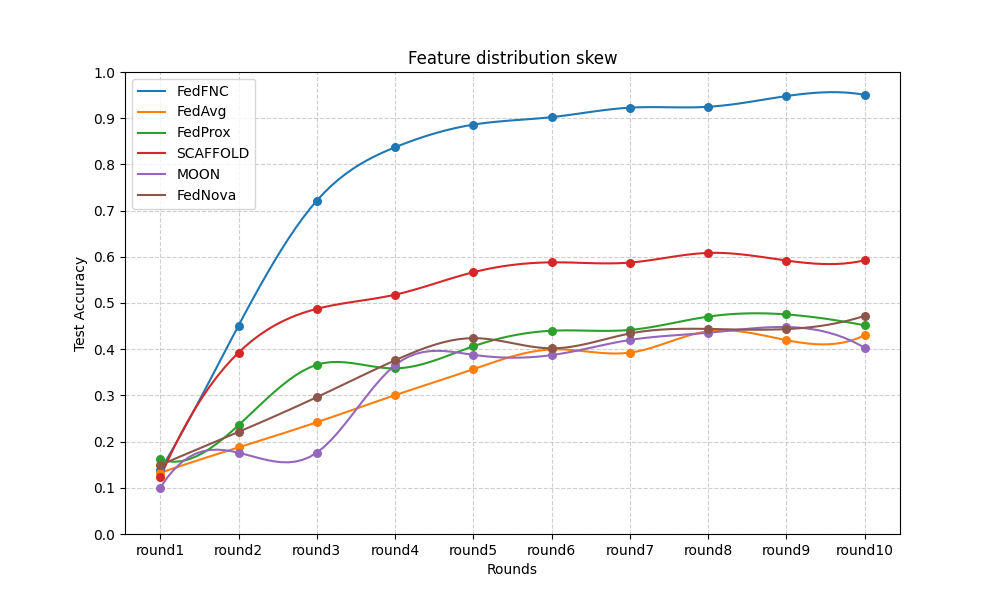}
		\caption{Feature distribution skew}
		\label{Covergence_FDS}
	\end{subfigure}
	\begin{subfigure}{0.3\textwidth}
		\includegraphics[width=\textwidth]{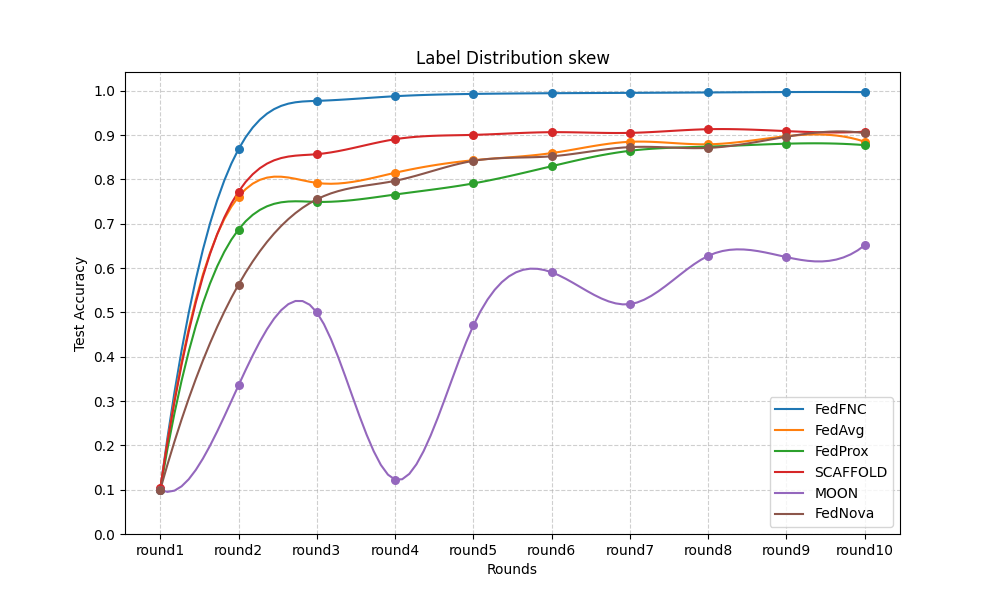}
		\caption{Label distribution skew}
		\label{Covergence_LDS}
	\end{subfigure}
	\begin{subfigure}{0.3\textwidth}
		\includegraphics[width=\textwidth]{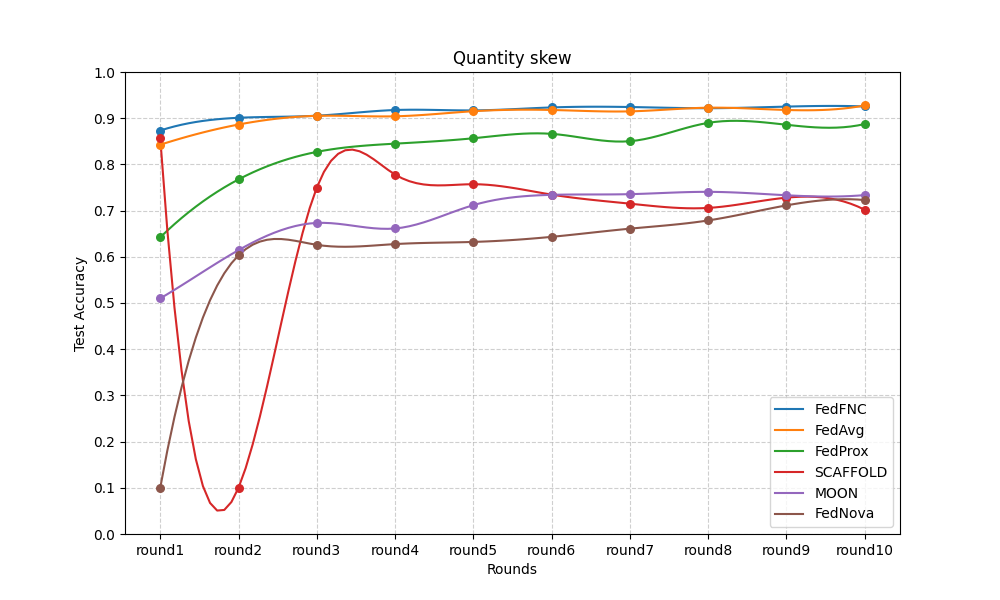}
		\caption{Quantity skew}
		\label{Covergence_QS}
	\end{subfigure}
	\caption{Convergence curves of training ResNet-18 on CIFAR-10}

	\label{Convergence}
\end{figure}

The convergence curves of training ResNet-18 on CIFAR-10 under three single types of non-i.i.d scenarios are illustrated in Figure \ref{Convergence}.
We can oberserve that the FNR-FL algorithm consistently exhibits the same level of convergence speed as other FL algorithms and achieves exceptionally better test accuracies across different non-i.i.d scenarios.
These results underline the potential of the FNR-FL algorithm to not only matchs the convergence speed of existing FL algorithms but also outperforms them in terms of the global model accuracy. 

\subsubsection{Communication and Computational Cost}

Metrics for performance evaluation are presented in Table \ref{tab:cost}, where the $Traffic$ represents the cumulative data exchanged, including both uploads and downloads, and $Time$ computes the duration to complete 10 rounds of training. 
The metrics $\kappa = Accuracy*{10^4} / Time$ and $\rho = Accuracy*{10^4} / Traffic$ were computed to measure test accuracy per unit of time and traffic, respectively, providing a quantifiable representation of  efficiency and data economy of each algorithm.
Among tested FL algorithms, FNR-FL stands out for its exceptional efficiency in non-i.i.d scenarios, such as a $\kappa$ of 1.6462 and a $\rho $ of 1.1184 in feature distribution skew. This overarching superiority of FNR-FL stems from its robust algorithmic structure that ensures high accuracy without disproportionately increasing communication or computational cost. 
  
\begin{table}[htbp]
	\centering
	\caption{Performance Evaluation of Federated Learning Algorithms (10 participants, 10 rounds)}
	  \begin{tabular}{@{}lccccccc@{}}
	  \toprule
	  & & \multicolumn{6}{c}{Federated Learning Algorithms} \\
	  \cmidrule{3-8}
	  Non-i.i.d scenarios & Metric & FNR-FL & FedAvg & FedProx & SCAFFOLD & MOON  & FedNova \\
	  \midrule
	  \multirow{4}{*}{Feature Distribution Skew} 
	  & Traffic (MB) & 8920  & 8920  & 8920  & 13380  & 8920  & 8920  \\
	  & Time (s) & 6060  & 6840  & 8160  & 6840  & 10692  & 6525  \\
	  & Accuracy & 0.9976  & 0.6001  & 0.5956  & 0.7425  & 0.5530  & 0.6223  \\
	  & $\kappa$ & \textbf{1.6462} & 0.8774  & 0.7299  & 1.0855  & 0.5172  & 0.9536  \\
	  & $\rho$ & \textbf{1.1184} & 0.6728  & 0.6677  & 0.5549  & 0.6199  & 0.6976  \\
	  \midrule
	  \multirow{4}{*}{Label Distribution Skew} 
	  & Traffic (MB) & 8920  & 8920  & 8920  & 13380  & 8920  & 8920  \\
	  & Time (s) & 7133  & 6006  & 11310  & 7176  & 8814  & 7098  \\
	  & Accuracy & 0.9970  & 0.8393  & 0.8773  & 0.9077  & 0.6515  & 0.9043  \\
	  & $\kappa$ & \textbf{1.3977} & 1.3974  & 0.7757  & 1.2649  & 0.7392  & 1.2740  \\
	  & $\rho$ & \textbf{1.1177} & 0.9409  & 0.9835  & 0.6784  & 0.7304  & 1.0138  \\
	  \midrule
	  \multirow{4}{*}{Quantity Skew} 
	  & Traffic (MB) & 8920  & 8920  & 8920  & 13380  & 8920  & 8920  \\
	  & Time (s) & 5400  & 6786  & 9360  & 7332  & 11544  & 6006  \\
	  & Accuracy & 0.9982  & 0.9107  & 0.8874  & 0.7017  & 0.7336  & 0.7293  \\
	  & $\kappa$ & \textbf{1.8485} & 1.3420  & 0.9481  & 0.9570  & 0.6355  & 1.2143  \\
	  & $\rho$ & \textbf{1.1191} & 1.0210  & 0.9948  & 0.5244  & 0.8224  & 0.8176  \\
	  \bottomrule
	  \end{tabular}%
	\label{tab:cost}%
  \end{table}%

\subsection{Performance Under Mixed Types of Non-i.i.d Scenarios}

In more realistic and challenging settings, where data distributions are often mixed, we evaluated the performance of federated learning algorithms on the CIFAR-10 dataset under combined scenarios of label and feature distribution skew as well as label and quantity distribution skew. Our findings are summarized in Table \ref{Exp2}.

\begin{enumerate}
	\item Label and Feature Distribution Skew:
	Under label and feature distribution skew with low noise levels (noise=0.1), FNR-FL achieves an impressive accuracy of 0.9755, significantly outperforming other FL algorithms including FedAvg, FedProx, SCAFFOLD, MOON, and FedNova. This performance differential becomes even more pronounced under high noise conditions (noise=0.5), where FNR-FL maintains its leading accuracy of 0.9111, while other FL algorithms experience a marked decline. This demonstrates FNR-FL's advanced capability to extract and generalize features effectively in mixed non-i.i.d scenarios.
	\item Label and Quantity Distribution Skew: 
	Under label and quantity distribution skew, a notable reduction in accuracy is observed across all algorithms, with FNR-FL maintaining the highest accuracy at 0.8826. The other FL algorithms demonstrate comparable results, with FedAvg, FedProx, SCAFFOLD, and FedNova clustering around the 0.3000 accuracy level, indicating a shared challenge in this mixed skew scenario.
\end{enumerate}

As noted by recent studies \cite{li2022federated} , mixed types of skew present more complex challenges than single types, often leading to a more pronounced degradation in model quality. FNR-FL's ability to maintain high accuracy levels under such conditions is indicative of its innovative design, addressing the critical need for effective algorithms capable of operating under mixed types of non-i.i.d scenarios, which is far more common in practical settings.

\begin{table}[htbp]
	\centering
	\caption{Test accuracy on CIFAR-10 under mixed types of non-i.i.d scenarios (ResNet-18)}
	  \begin{tabular}{cccccccc}
	  \toprule
	  \multicolumn{2}{c}{Non-i.i.d scenarios} & FNR-FL & FedAvg & FedProx & SCAFFOLD & MOON  & FedNova \\
	  \midrule 
	  \multirow{2}[0]{*}{label+feature skew} & noise=0.1 & 0.9755 & 0.80665 & 0.78365 & 0.83705 & 0.636 & 0.8029 \\
			& noise=0.5 & 0.9111 & 0.4504 & 0.4587 & 0.5763 & 0.37275 & 0.3942 \\
	  \multicolumn{2}{c}{label+quantity skew} & 0.8826 & 0.309439 & 0.304785 & 0.311153 & 0.2993 & 0.309439 \\
	  \bottomrule
	  \end{tabular}%
	\label{Exp2}%
  \end{table}%

\subsection{Orthogonality of FNR-FL with existing FL algorithms}

The experimental results in Table \ref{Exp_or} demonstrate the effectiveness of integrating the FNR-FL algorithm with other FL algorithms.
Across all scenarios, combinations involving FNR-FL consistently outperform their standalone FL algorithms, indicating a substantial enhancement in handling non-i.i.d data distributions. Specifically, FedAvg+FNR-FL emerges as a top performer in feature and label distribution skews, achieving accuracies of 0.9976 and 0.9970 respectively, while FedProx+FNR-FL leads in the quantity skew scenario with an accuracy of 0.9999. These results suggest that FNR-FL's regularization mechanism effectively complements and augments existing FL algorithms \cite{acar2021federated}.

\begin{table}[htbp]
	\centering
	\caption{Orthogonality of FNR-FL with existing FL algorithms (ResNet-18, CIFAR-10)}
	  \begin{tabular}{cccc}
	  \toprule
	  \multicolumn{1}{c}{Non-i.i.d scenarios} & \multicolumn{1}{c}{Feature distribution skew} & \multicolumn{1}{c}{Label distribution skew} & \multicolumn{1}{c}{Quantity skew} \\
	  \midrule 
	  FedAvg+FNR-FL & \textbf{0.9976 } & \textbf{0.9970 } & 0.9982  \\
	  FedProx+FNR-FL & 0.9784  & 0.9838  & \textbf{0.9999 } \\
	  SCAFFOLD+FNR-FL & 0.9783  & 0.9892  & 0.9637  \\
	  MOON+FNR-FL & 0.9678  & 0.9774  & 0.9532  \\
	  FedNova+FNR-FL & 0.9664  & 0.9766  & 0.9427  \\
	  \bottomrule
	  \end{tabular}%
	\label{Exp_or}%
  \end{table}%

\subsection{Effect of feature norm regularization}

The ablation study assesses the impact of feature norm regularization (FNR) within the FNR-FL algorithm on a participant's local model. 
Notably, classes with lower local sample sizes, such as 'frog' (471) and 'horse' (485), show more significant accuracy improvements. This observation is aligned with the hypothesis that FNR enables participants to benefit from the shared knowledge across the federation, particularly when their local data is insufficient to capture the complexity of the class.
In contrast, classes with more substantial local representations, like 'automobile' (512) and 'ship' (528), exhibit smaller gains. This can be attributed to the participant's local model already achieving a higher accuracy due to the ample class-specific data, hence the marginal utility of the shared knowledge from FNR is less impactful \cite{du2022rethinking, shao2023survey, qi2023differentially}.
These findings underscore FNR's role in mitigating the decrease of test accuracy brought by non-i.i.d data distribution, contributing to the robustness of the FNR-FL algorithm.

\begin{figure}
	\centering
	\includegraphics[width=0.6\textwidth]{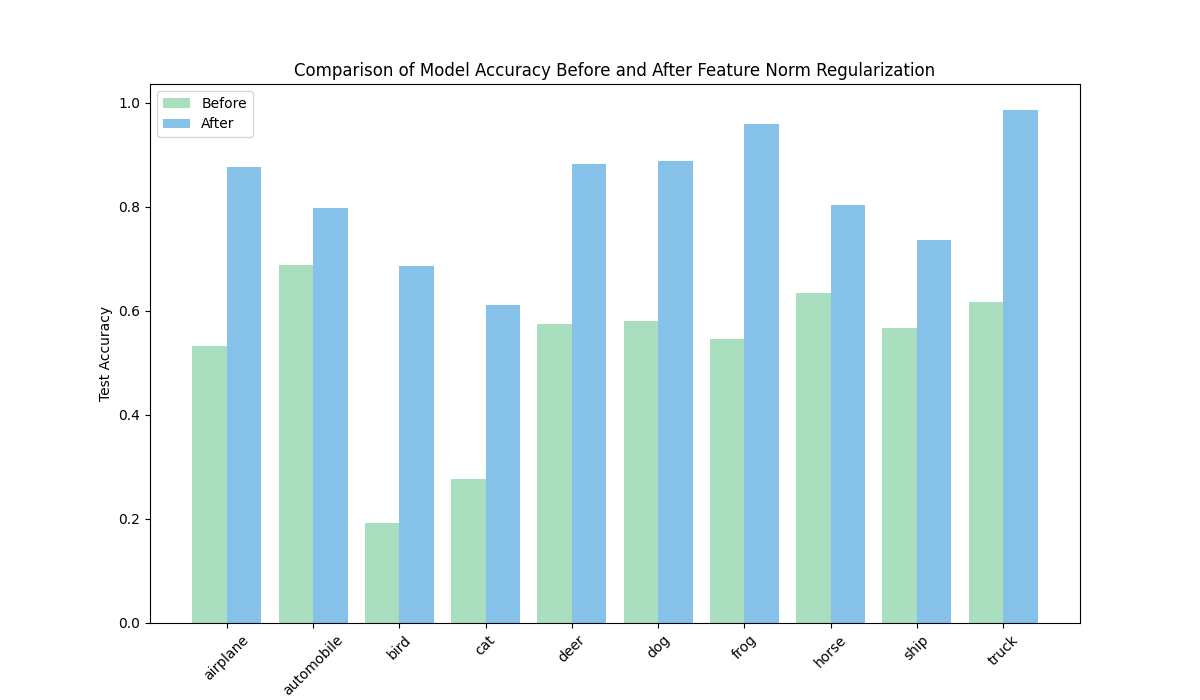} 
	\caption{Comparison of Model Accuracy Before and After Feature Norm Regularization}
	\label{effectofFNR}
\end{figure}

\subsection{Effect of noise}

The experiment evaluates the performance and robustness of FNR-FL against other FL algorithms across different noise levels within feature and label distribution skews. The results are presented in Table \ref{Exp_noise}.
In the feature distribution skew, FNR-FL consistently maintains the highest accuracy, indicating a strong resilience to noise. With a standard deviation (SD) of 0.0 (no noise), FNR-FL achieves near-perfect accuracy (0.9995), significantly outperforming the next best algorithm, SCAFFOLD, by a margin of 7\%. As noise increases (SDs of 0.1 and 0.5), FNR-FL notably sustains a lesser degree of decline relative to other algorithms. 
A similar pattern is observed in label distribution skew scenarios. 
These findings suggest that FNR-FL is robust against noise in the data. This resilience to noise highlights the algorithm's potential for practical deployment in federated learning scenarios \cite{wu2023fednoro}.

\begin{table}[htbp]
	\centering
	\caption{Effect of noise (ResNet-18, CIFAR-10)}
	  \begin{tabular}{cccccccc}
	  \toprule
	  \multicolumn{1}{c}{Non-i.i.d setting} & Noise & FNR-FL & FedAvg & FedProx & SCAFFOLD & MOON  & FedNova \\
	  \midrule 
	  \multicolumn{1}{c}{\multirow{3}[0]{*}{Feature distribution skew}} & 0.0   & \textbf{0.9995 } & 0.9238  & 0.9039  & 0.9290  & 0.7371  & 0.9257  \\
			& 0.1   & \textbf{0.9984 } & 0.8549  & 0.8219  & 0.8581  & 0.5677  & 0.8469  \\
			& 0.5   & \textbf{0.9505 } & 0.4303  & 0.4517  & 0.5925  & 0.4030  & 0.4723  \\
	  \multicolumn{1}{c}{\multirow{3}[0]{*}{Label distribution skew}} & 0.0   & \textbf{0.9970 } & 0.8850  & 0.8773  & 0.9077  & 0.6515  & 0.9043  \\
			& 0.1   & \textbf{0.9755 } & 0.8067  & 0.7837  & 0.8371  & 0.6360  & 0.8029  \\
			& 0.5   & \textbf{0.9111 } & 0.4504  & 0.4587  & 0.5763  & 0.3728  & 0.3942  \\
	  \bottomrule
	  \end{tabular}%
	\label{Exp_noise}%
  \end{table}%

\section{Conclusions and Future Work}

In this paper, the Feature Norm Regularization Federated Learning (FNR-FL) algorithm is proposed as a novel approach to enhancing performance in non-i.i.d data scenarios. The core philosophy of FNR-FL revolves around leveraging feature norms as a metric to quantify and subsequently mitigate the divergence in local model updates. Experimental evaluations have shown that FNR-FL achieves superior test accuracy without loss of convergence and excessive computational and communication overhead under single types of non-i.i.d scenarios. For mixed non-i.i.d scenarios, FNR-FL has proven to address complex data challenges with the highest recorded accuracy, marking the first known federated learning algorithm tested under such mixed conditions. The orthogonal experiment underscores FNR-FL's modularity and plug-and-play nature, seamlessly enhancing existing FL algorithms to deliver marked improvements. The necessity of feature norm regularization is validated through the ablation experiment. Moreover, FNR-FL maintains robustness against different levels of noise added in samples.

For future work, we aim to investigate the optimization of FNR-FL for further communication cost reduction through combination with model compression methods. Additionally, we plan to explore the application of FNR-FL to privacy-preserving federated learning, broadening its scope to prove its versatility and efficacy in various domains. These initiatives will allow us to refine FNR-FL's capabilities, ensuring its adaptability and performance in increasingly complex federated learning environments. Furthermore, we will extend our research to develope a decision-making algorithm capable of analyzing the characteristics of data distribution in federated learning and subsequently recommending the most suitable FL algorithm for the task at hand.

\bibliographystyle{unsrtnat}
\bibliography{references}

\begin{appendices}

\section{Proof of convergence analysis}
\label{Proof}

\subsection{Assumptions}

We make the following assumptions in our analysis:

\begin{itemize}
    \item[\textbf{1.}] The functions $\Phi_1, \ldots, \Phi_N$ are all $L$-smooth. This implies that for all $x, y$, the following inequality holds:
    \begin{equation}
        \Phi_i(y) \leq \Phi_i(x) + \langle \nabla\Phi_i(x), y-x \rangle + \frac{L}{2} \| y-x \|^2.
    \end{equation}
    
    \item[\textbf{2.}] The functions are all $\mu$-strongly convex, which means for all $x, y$:
    \begin{equation}
        \Phi_i(y) \geq \Phi_i(x) + \langle \nabla\Phi_i(x), y-x \rangle + \frac{\mu}{2} \| y-x \|^2.
    \end{equation}
    
    \item[\textbf{3.}] The stochastic gradients are unbiased and have bounded variance. Specifically, in $m$-th mini-batch gradient descent step of participant $P_i$:
    \begin{equation}
        \mathbb{E} \left[ \| \nabla \Phi_i(w^i_{m}, b^i_{m}) - \nabla \Phi_i(w^i_{m}) \|^2 \right] \leq ({\Delta G}_i)^2,
    \end{equation}
    and
    \begin{equation}
        \mathbb{E} \left[ \| \nabla \Phi_i(w^i_{m}, b^i_{m}) \|^2 \right] \leq G^2.
    \end{equation}

	where $b^i_m$ denotes a random batch in all batches of participant $P_i$ in $m$-th mini-batch gradient descent step.
\end{itemize}

\subsection{Proof}

Consider the $m$-th mini-batch gradient descent step: $\bar{\mathbf{w}}_{m+1} = \bar{\mathbf{w}}_m - \eta_m \mathbf{g}_m$. We analyze the distance to the optimal point $\mathbf{w}^*$ in terms of the squared norm:
\begin{equation}
	\begin{aligned}
		\|\bar{\mathbf{w}}_{m+1} - \mathbf{w}^*\|^2 &= \|\bar{\mathbf{w}}_m - \eta_m \mathbf{g}_m - \mathbf{w}^*\|^2 \\
		&= \|\bar{\mathbf{w}}_m - \eta_m \mathbf{g}_m - \mathbf{w}^* - \eta_m \bar{\mathbf{g}}_m + \eta_m \bar{\mathbf{g}}_m\|^2 \\
		&= \|(\bar{\mathbf{w}}_m - \mathbf{w}^* - \eta_m \bar{\mathbf{g}}_m )+ \eta_m (\bar{\mathbf{g}}_m - \mathbf{g}_m)\|^2 \\
		&= \|\bar{\mathbf{w}}_m  -\mathbf{w}^* - \eta_m \bar{\mathbf{g}}_m \|^2 + \eta_m^2 \|\bar{\mathbf{g}}_m - \mathbf{g}_m\|^2 \\
		&\quad + 2\eta_m \langle \bar{\mathbf{w}}_m - \mathbf{w}^* - \eta_m \bar{\mathbf{g}}_m, \bar{\mathbf{g}}_m - \mathbf{g}_m \rangle.
	\end{aligned}
	\label{i}
\end{equation}

where $\mathbf{g}_m = \nabla\Phi_i(\mathbf{w}^i_m, b^i_{m})$, and $\bar{\mathbf{g}}_m = \nabla\Phi_i(\mathbf{w}^i_m)$.

\subsubsection{Bounding $\|\bar{\mathbf{w}}_m  -\mathbf{w}^* - \eta_m \bar{\mathbf{g}}_m \|^2$}

Expanding the squared norm, we have:
\begin{equation}
	\begin{aligned}
		\|\bar{\mathbf{w}}_{m} - \mathbf{w}^* - \eta_m \bar{\mathbf{g}}_m\|^2 &= \|\bar{\mathbf{w}}_{m} - \mathbf{w}^*\|^2 - 2\eta_m \langle \bar{\mathbf{w}}_{m} - \mathbf{w}^*, \mathbf{g}_m \rangle + \eta_m^2 \|\mathbf{g}_m\|^2.  
	\end{aligned}
\end{equation}

Letting $ \mathbf{H}  = 2\eta_m \langle \bar{\mathbf{w}}_{m} - \mathbf{w}^*, \mathbf{g}_m \rangle$, $\mathbf{I} = \eta_m^2 \|\mathbf{g}_m\|^2$.

\paragraph{Bounding term $\mathbf{I}$ }

For the term $\mathbf{I}$, by the definition of $L$-smoothness of $\Phi_i$, we have:
\begin{equation}
\Phi_i(y) \leq \Phi_i(x) + \langle \nabla\Phi_i(x), y-x \rangle + \frac{L}{2} \| y-x \|_2^2,
\end{equation}
for any $x$ and $y$. Choosing $y = \mathbf{w}$ and $x = \mathbf{w}^*$, and since $\nabla\Phi_i(\mathbf{w}^*) = 0$ by the optimality of $\mathbf{w}^*$, we get:
\begin{equation}
\Phi_i(\mathbf{w}) - \Phi_i(\mathbf{w}^*) \leq \frac{L}{2} \| \mathbf{w} - \mathbf{w}^*\|_2^2.
\label{a}
\end{equation}

According to the definition of $L$-smoothness, the gradient norm can be bounded as:
\begin{equation}
\| \nabla\Phi_i(\mathbf{w}) - \nabla\Phi_i(\mathbf{w}^*) \| \leq L \| \mathbf{w} - \mathbf{w}^*\|.
\end{equation}

Since $\nabla\Phi_i(\mathbf{w}^*) = 0$, it follows that:
\begin{equation}
\| \nabla\Phi_i(\mathbf{w}) \| \leq L \| \mathbf{w} - \mathbf{w}^*\|. 
\end{equation}

And therefore:
\begin{equation}
\| \nabla\Phi_i(\mathbf{w}) \|^2 \leq L^2 \| \mathbf{w} - \mathbf{w}^*\|^2.
\label{b}
\end{equation}

Combining \eqref{a} and \eqref{b}, we get:
\begin{equation}
\| \nabla\Phi_i(\mathbf{w}) \|^2 \leq 2L (\Phi_i(\mathbf{w}) - \Phi_i(\mathbf{w}^*)).
\end{equation}

We define the weighted average gradient $\bar{\mathbf{g}}_m$ among participants $P_i, \cdots, P_n$ as:
\begin{equation}
	\bar{\mathbf{g}}_m = \sum_{i=1}^{n} {(\frac{|D^{i}|}{\sum^{n}_{i=1}|D^{i}|})} \nabla\Phi_i(\mathbf{w}^i_m),
\label{barg}
\end{equation}
and consequently,
\begin{equation}
\|\bar{\mathbf{g}}_m\|^2 = \left\| \sum_{i=1}^{n} {(\frac{|D^{i}|}{\sum^{n}_{i=1}|D^{i}|})} \nabla\Phi_i(\mathbf{w}^i_m) \right\|^2,
\end{equation}
which leads to
\begin{equation}
	\begin{aligned}
		\MoveEqLeft \|\bar{\mathbf{g}}_m\|^2 = \left\| \sum_{i=1}^{n} {(\frac{|D^{i}|}{\sum^{n}_{i=1}|D^{i}|})} \nabla\Phi_i(\mathbf{w}^i_m) \right\|^2 \\
		&= \sum_{i=1}^{n} {(\frac{|D^{i}|}{\sum^{n}_{i=1}|D^{i}|})}^2 \|\nabla\Phi_i(\mathbf{w}^i_m)\|^2 + \sum_{i=1}^{n} \sum_{j \neq k} p_k p_j \langle \nabla\Phi_i(\mathbf{w}^k_m), \nabla\Phi_i(\mathbf{w}^j_m) \rangle.
	\end{aligned}
\end{equation}

Applying the Cauchy-Schwarz inequality, we get
\begin{equation}
\sum_{i=1}^{n} \sum_{j \neq k} p_k p_j \langle \nabla\Phi_i(\mathbf{w}^k_m), \nabla\Phi_j(\mathbf{w}^j_m) \rangle \leq \sum_{i=1}^{n} {(\frac{|D^{i}|}{\sum^{n}_{i=1}|D^{i}|})} \|\nabla\Phi_i(\mathbf{w}^i_m)\|^2.
\end{equation}

Finally, for term $\mathbf{I}$ we have:
\begin{equation}
\eta_m^2 \|\bar{\mathbf{g}}_m\|^2 \leq \eta_m^2 \sum_{i=1}^{n} {(\frac{|D^{i}|}{\sum^{n}_{i=1}|D^{i}|})} \|\nabla\Phi_i(\mathbf{w}^i_m)\|^2.
\end{equation}

\paragraph{Bounding term $\mathbf{H}$ }

For term $\mathbf{H}$, we have:
\begin{align}
\langle \bar{\mathbf{w}}_{m} - \mathbf{w}^*, \bar{\mathbf{g}}_m \rangle 
&= \langle \bar{\mathbf{w}}_{m} - \mathbf{w}^* + \mathbf{w}^i_{m} - \mathbf{w}^i_{m}, \bar{\mathbf{g}}_m \rangle \nonumber \\
&= \langle \bar{\mathbf{w}}_{m} - \mathbf{w}^i_{m} , \bar{\mathbf{g}}_m \rangle + \langle \mathbf{w}^i_{m} - \mathbf{w}^*, \bar{\mathbf{g}}_m \rangle \nonumber \\
&= \|\bar{\mathbf{w}}_{m} - \mathbf{w}^i_{m}\| \|\bar{\mathbf{g}}_m\| \cos \theta + \langle \mathbf{w}^i_{m} - \mathbf{w}^*, \bar{\mathbf{g}}_m \rangle \nonumber \\
&\geq \|\bar{\mathbf{w}}_{m} - \mathbf{w}^i_{m}\| \|\bar{\mathbf{g}}_m\| + \langle \mathbf{w}^i_{m} - \mathbf{w}^*, \bar{\mathbf{g}}_m \rangle \nonumber \\
&= -\sum_{i=1}^{n} {(\frac{|D^{i}|}{\sum^{n}_{i=1}|D^{i}|})} \|\bar{\mathbf{w}}_{m} - \mathbf{w}^i_{m}\| \|\nabla\Phi_i(\mathbf{w}^i_{m})\| + \langle \mathbf{w}^i_{m} - \mathbf{w}^*, \bar{\mathbf{g}}_m \rangle \nonumber \\
&\geq -\frac{1}{2} \sum_{i=1}^{n} {(\frac{|D^{i}|}{\sum^{n}_{i=1}|D^{i}|})} \left( \frac{1}{\eta_m} \|\bar{\mathbf{w}}_{m} - \mathbf{w}^i_{m}\|^2 + \eta_m \|\nabla\Phi_i(\mathbf{w}^i_{m})\|^2 \right) + \langle \mathbf{w}^i_{m} - \mathbf{w}^*, \bar{\mathbf{g}}_m \rangle.
\label{d}
\end{align}

According to $\mu$-strongly convex:
\begin{equation}
\Phi_i(y) \geq \Phi_i(x) + \langle y - x, \nabla\Phi_i(x) \rangle + \frac{\mu}{2} \|y - x\|_2^2
\end{equation}
Letting $y = \mathbf{w}^*$ and $x = \mathbf{w}^i_m$, we have:
\begin{equation}
\Phi_i(\mathbf{w}^*) \geq \Phi_i(\mathbf{w}^i_m) - \langle \mathbf{w}^* - \mathbf{w}^i_m, \nabla\Phi_i(\mathbf{w}^i_m) \rangle + \frac{\mu}{2} \|\mathbf{w}^* - \mathbf{w}^i_m\|_2^2
\label{c}
\end{equation}

According to \eqref{barg}, we have: 
\begin{equation}
\sum_{i=1}^{n} {(\frac{|D^{i}|}{\sum^{n}_{i=1}|D^{i}|})} \langle \mathbf{w}^* - \mathbf{w}^i_m, \nabla\Phi_i(\mathbf{w}^i_m) \rangle = \langle \mathbf{w}^* - \mathbf{w}^i_m, \bar{\mathbf{g}}_m \rangle
\end{equation}

According to \eqref{c}, we have: 
\begin{equation}
\langle \bar{\mathbf{w}}_m - \mathbf{w}^*, \nabla\Phi_i(\bar{\mathbf{w}}_m) \rangle \geq \Phi_i(\mathbf{w}^*) - \Phi_i(\bar{\mathbf{w}}_m) + \frac{\mu}{2} \|\mathbf{w}^* - \bar{\mathbf{w}}_m\|_2^2
\end{equation}

Continuing from \eqref{d}, we combine \eqref{barg} and \eqref{c}, thus getting:
\begin{equation}
	\begin{aligned}
		\langle \bar{\mathbf{w}}_{m} - \mathbf{w}^*, \bar{\mathbf{g}}_m \rangle  
		&\geq -\frac{1}{2} \sum_{i=1}^{n} {(\frac{|D^{i}|}{\sum^{n}_{i=1}|D^{i}|})} \left( \frac{1}{\eta_m} \|\bar{\mathbf{w}}_{m} - \mathbf{w}^i_{m}\|^2 + \eta_m \|\nabla\Phi_i(\mathbf{w}^i_{m})\|^2 \right) \\
		&\quad + \sum_{i=1}^{n} {(\frac{|D^{i}|}{\sum^{n}_{i=1}|D^{i}|})} \bigg[\Phi_i(\mathbf{w}^*) - \Phi_i(\bar{\mathbf{w}}_m) + \frac{\mu}{2} \|\mathbf{w}^* - \bar{\mathbf{w}}_m\|_2^2 \bigg].
	\end{aligned}
\end{equation}

And therefore: 

\begin{equation}
	\begin{aligned}
		\|\bar{\mathbf{w}}_m - \mathbf{w}^* - \eta_m \bar{\mathbf{g}}_m\|^2 &= \|\bar{\mathbf{w}}_m - \mathbf{w}^*\|^2 - 2\eta_m \langle \bar{\mathbf{w}}_m - \mathbf{w}^*, \bar{\mathbf{g}}_m \rangle + \eta_m^2 \|\bar{\mathbf{g}}_m\|^2 \\
		&\leq \|\bar{\mathbf{w}}_m - \mathbf{w}^*\|^2 - 2\eta_m \sum_{i=1}^{n} {(\frac{|D^{i}|}{\sum^{n}_{i=1}|D^{i}|})} \bigg[\Phi_i(\mathbf{w}^*) - \Phi_i(\bar{\mathbf{w}}_m) + \frac{\mu}{2} \|\mathbf{w}^* - \bar{\mathbf{w}}_m\|_2^2 \bigg]	\\
		&\quad + \eta_m \sum_{i=1}^{n} {(\frac{|D^{i}|}{\sum^{n}_{i=1}|D^{i}|})} \left( \frac{1}{\eta_m} \|\bar{\mathbf{w}}_{m} - \mathbf{w}^i_{m}\|^2 + \eta_m \|\nabla\Phi_i(\mathbf{w}^i_{m})\|^2 \right) + \\
		&\eta_m^2 \sum_{i=1}^{n} {(\frac{|D^{i}|}{\sum^{n}_{i=1}|D^{i}|})} \|\nabla\Phi_i(\mathbf{w}^i_m)\|^2 \\
		&\leq \|\bar{\mathbf{w}}_m - \mathbf{w}^*\|^2 - 2\eta_m \sum_{i=1}^{n} {(\frac{|D^{i}|}{\sum^{n}_{i=1}|D^{i}|})} (\Phi_i(\mathbf{w}^i_m) - \Phi_i(\mathbf{w}^*)) \\
		&\quad - \mu\eta_m \sum_{i=1}^{n} {(\frac{|D^{i}|}{\sum^{n}_{i=1}|D^{i}|})} \|\mathbf{w}^* - \bar{\mathbf{w}}_m\|_2^2 + 4L\eta_m^2 \sum_{i=1}^{n} {(\frac{|D^{i}|}{\sum^{n}_{i=1}|D^{i}|})} (\Phi_i(\mathbf{w}^i_m) - \Phi_i(\mathbf{w}_i^*)) \\
		&= (1-\mu\eta_m) \|\bar{\mathbf{w}}_m - \mathbf{w}^*\|^2 + \sum_{i=1}^{n} {(\frac{|D^{i}|}{\sum^{n}_{i=1}|D^{i}|})} \|\bar{\mathbf{w}}_m - \mathbf{w}^i_m\|_2^2 \\
		&\quad + 4L\eta_m^2 \sum_{i=1}^{n} {(\frac{|D^{i}|}{\sum^{n}_{i=1}|D^{i}|})} (\Phi_i(\mathbf{w}^i_m) - \Phi_i^*) - 2\eta_m \sum_{i=1}^{n} {(\frac{|D^{i}|}{\sum^{n}_{i=1}|D^{i}|})} (\Phi_i(\mathbf{w}^i_m) - \Phi_i(\mathbf{w}^*)). 
	\end{aligned}
	\label{e}
\end{equation}

Letting $\mathbf{J} = 4L\eta_m^2 \sum_{i=1}^{n} {(\frac{|D^{i}|}{\sum^{n}_{i=1}|D^{i}|})} (\Phi_i(\mathbf{w}^i_m) - \Phi_i^*) - 2\eta_m \sum_{i=1}^{n} {(\frac{|D^{i}|}{\sum^{n}_{i=1}|D^{i}|})} (\Phi_i(\mathbf{w}^i_m) - \Phi_i(\mathbf{w}^*))$.

\paragraph{Bounding term $\mathbf{J} $}

\begin{equation}
	\begin{aligned}
		\MoveEqLeft 4L\eta_m^2 \sum_{i=1}^{n} {(\frac{|D^{i}|}{\sum^{n}_{i=1}|D^{i}|})} (\Phi_i(\mathbf{w}^i_m) - \Phi_i^*) - 2\eta_m \sum_{i=1}^{n} {(\frac{|D^{i}|}{\sum^{n}_{i=1}|D^{i}|})} (\Phi_i(\mathbf{w}^i_m) - \Phi_i(\mathbf{w}^*) + \Phi_i^* - \Phi_i^* ) \\
		&= 4L\eta_m^2 \sum_{i=1}^{n} {(\frac{|D^{i}|}{\sum^{n}_{i=1}|D^{i}|})} (\Phi_i(\mathbf{w}^i_m) - \Phi_i^*) - 2\eta_m \sum_{i=1}^{n} {(\frac{|D^{i}|}{\sum^{n}_{i=1}|D^{i}|})} (\Phi_i(\mathbf{w}^i_m) - \Phi_i^*) \\ 
		&\quad + 2\eta_m \sum_{i=1}^{n} {(\frac{|D^{i}|}{\sum^{n}_{i=1}|D^{i}|})} (\Phi_i(\mathbf{w}^*) - \Phi_i^*) \\
			&= 2\eta_m (2L\eta_m - 1) \sum_{i=1}^{n} {(\frac{|D^{i}|}{\sum^{n}_{i=1}|D^{i}|})} (\Phi_i(\mathbf{w}^i_m) - \Phi_i^*) + 2\eta_m \sum_{i=1}^{n} {(\frac{|D^{i}|}{\sum^{n}_{i=1}|D^{i}|})} (\Phi_i(\mathbf{w}^*) - \Phi_i^*) \\
			&= 2\eta_m (2L\eta_m - 1) \sum_{i=1}^{n} {(\frac{|D^{i}|}{\sum^{n}_{i=1}|D^{i}|})} [(\Phi_i(\mathbf{w}^i_m) - \Phi^*) - (\Phi_i^* - \Phi^*)] + 2\eta_m \sum_{i=1}^{n} {(\frac{|D^{i}|}{\sum^{n}_{i=1}|D^{i}|})} (\Phi_i(\mathbf{w}^*) - \Phi_i^*) \\
			&= 2\eta_m (2L\eta_m - 1) \sum_{i=1}^{n} {(\frac{|D^{i}|}{\sum^{n}_{i=1}|D^{i}|})} (\Phi_i(\mathbf{w}^i_m) - \Phi^*) + \left[2\eta_m (2L\eta_m - 1) + 2\eta_m\right] \Gamma  \\
			&= 2\eta_m (2L\eta_m - 1) \sum_{i=1}^{n} {(\frac{|D^{i}|}{\sum^{n}_{i=1}|D^{i}|})} (\Phi_i(\mathbf{w}^i_m) - \Phi^*) + 4L\eta_m^2 \Gamma.
	\end{aligned}
\end{equation}

where $\Gamma = \sum_{i=1}^{n} {(\frac{|D^{i}|}{\sum^{n}_{i=1}|D^{i}|})} (\Phi^* - \Phi_i^*) = \Phi^* - \sum_{i=1}^{n} {(\frac{|D^{i}|}{\sum^{n}_{i=1}|D^{i}|})} \Phi_i^* = \Phi(\mathbf{w}^*) - \sum_{i=1}^{n} {(\frac{|D^{i}|}{\sum^{n}_{i=1}|D^{i}|})} \Phi_i(\mathbf{w}^*)$.

Letting $\mathbf{K} = 2\eta_m (2L\eta_m - 1) \sum_{i=1}^{n} {(\frac{|D^{i}|}{\sum^{n}_{i=1}|D^{i}|})} (\Phi_i(\mathbf{w}^i_m) - \Phi^*)$.

\paragraph{Bounding term $\mathbf{K} $}

According to $\mu$-strongly convex:
\begin{equation}
\Phi_i(y) \geq \Phi_i(x) + \langle y - x, \nabla\Phi_i(x) \rangle + \frac{\mu}{2} \|y - x\|_2^2
\end{equation}
Let $y = \mathbf{w}^i_m$ and $x = \bar{\mathbf{w}}_m$, we have:
\begin{equation}
\Phi_i(\bar{\mathbf{w}}_m) - \Phi_i(\mathbf{w}^i_m) \leq \langle \bar{\mathbf{w}}_m - \mathbf{w}^i_m, \nabla\Phi_i(\bar{\mathbf{w}}_m) \rangle - \frac{\mu}{2} \|\bar{\mathbf{w}}_m - \mathbf{w}^i_m\|_2^2 \leq \langle \bar{\mathbf{w}}_m - \mathbf{w}^i_m, \nabla\Phi_i(\bar{\mathbf{w}}_m) \rangle
\end{equation}

For term $\mathbf{K} $, we have:
\begin{equation}
	\begin{aligned}
		\MoveEqLeft \sum_{i=1}^{n} {(\frac{|D^{i}|}{\sum^{n}_{i=1}|D^{i}|})} (\Phi_i(\mathbf{w}^i_m) - \Phi^*) \\
		&= \sum_{i=1}^{n} {(\frac{|D^{i}|}{\sum^{n}_{i=1}|D^{i}|})} (\Phi_i(\mathbf{w}^i_m) - \Phi_i(\bar{\mathbf{w}}_m)) + \sum_{i=1}^{n} {(\frac{|D^{i}|}{\sum^{n}_{i=1}|D^{i}|})} (\Phi_i(\bar{\mathbf{w}}_m) - \Phi^*) \\
		&\geq \sum_{i=1}^{n} {(\frac{|D^{i}|}{\sum^{n}_{i=1}|D^{i}|})} \langle \mathbf{w}^i_m - \bar{\mathbf{w}}_m, \nabla\Phi_i(\bar{\mathbf{w}}_m) \rangle + \sum_{i=1}^{n} {(\frac{|D^{i}|}{\sum^{n}_{i=1}|D^{i}|})} (\Phi_i(\bar{\mathbf{w}}_m) - \Phi^*) \\
		&\geq -\frac{1}{2} \sum_{i=1}^{n} {(\frac{|D^{i}|}{\sum^{n}_{i=1}|D^{i}|})} \left[ \frac{1}{\eta_m} \|\mathbf{w}^i_m - \bar{\mathbf{w}}_m\|^2 + \eta_m \|\nabla\Phi_i(\bar{\mathbf{w}}_m)\|^2 \right] + (\Phi(\bar{\mathbf{w}}_m) - \Phi^*) \\
		&\geq -\sum_{i=1}^{n} {(\frac{|D^{i}|}{\sum^{n}_{i=1}|D^{i}|})} \left[ \frac{1}{2\eta_m} \|\mathbf{w}^i_m - \bar{\mathbf{w}}_m\|^2 + \eta_m L \|\Phi_i(\bar{\mathbf{w}}_m) - \Phi_i^*\|^2 \right] + (\Phi(\bar{\mathbf{w}}_m) - \Phi^*).
	\end{aligned}		
\end{equation}

Therefore, for term $\mathbf{J} $, we have: 
\begin{equation}
	\begin{aligned}
		\MoveEqLeft 4L\eta_m^2 \sum_{i=1}^{n} {(\frac{|D^{i}|}{\sum^{n}_{i=1}|D^{i}|})} (\Phi_i(\mathbf{w}^i_m) - \Phi_i^*) - 2\eta_m \sum_{i=1}^{n} {(\frac{|D^{i}|}{\sum^{n}_{i=1}|D^{i}|})} (\Phi_i(\mathbf{w}^i_m) - \Phi_i(\mathbf{w}^*))	\\
		&= 2\eta_m (1 - 2L\eta_m) \sum_{i=1}^{n} {(\frac{|D^{i}|}{\sum^{n}_{i=1}|D^{i}|})} \left[ \frac{1}{2\eta_m} \| \mathbf{w}^i_m - \bar{\mathbf{w}}_m\|^2 + \eta_m L(\Phi_i(\bar{\mathbf{w}}_m) - \Phi_i^*) \right]\\
		&\quad - 2\eta_m (1 - 2L\eta_m) (\Phi_i(\bar{\mathbf{w}}_m) - \Phi^*) + 4L\eta_m^2 \Gamma \\
		&= (1 - 2L\eta_m) \sum_{i=1}^{n} {(\frac{|D^{i}|}{\sum^{n}_{i=1}|D^{i}|})} \|\mathbf{w}^i_m - \bar{\mathbf{w}}_m\|^2 + 2{\eta_m}^2 (1 - 2L\eta_m) \sum_{i=1}^{n} {(\frac{|D^{i}|}{\sum^{n}_{i=1}|D^{i}|})} \left[ (\Phi_i(\bar{\mathbf{w}}_m) - \Phi^*) - (\Phi_i^* - \Phi^*) \right] \\
		&\quad - 2\eta_m (1 - 2L\eta_m) (\Phi(\bar{\mathbf{w}}_m) - \Phi^*) + 4L\eta_m^2 \Gamma \\
		&= A \sum_{i=1}^{n} {(\frac{|D^{i}|}{\sum^{n}_{i=1}|D^{i}|})} \|\mathbf{w}^i_m - \bar{\mathbf{w}}_m\|^2 + 2\eta_m A \sum_{i=1}^{n} {(\frac{|D^{i}|}{\sum^{n}_{i=1}|D^{i}|})} (\Phi_i(\bar{\mathbf{w}}_m) - \Phi^*) \\
		&\quad + 2L\eta_m^2 A \Gamma - 2\eta_m A (\Phi(\bar{\mathbf{w}}_m) - \Phi_i^*) + 4L\eta_m^2 \Gamma \\
		&= A \sum_{i=1}^{n} {(\frac{|D^{i}|}{\sum^{n}_{i=1}|D^{i}|})} \|\mathbf{w}^i_m - \bar{\mathbf{w}}_m\|^2 + 2 L {\eta_m}^2 (2 + A) \Gamma  + 2\eta_m A (\eta_m L - 1) \sum_{i=1}^{n} {(\frac{|D^{i}|}{\sum^{n}_{i=1}|D^{i}|})} (\Phi_i(\bar{\mathbf{w}}_m) - \Phi^*)
	\end{aligned}
\end{equation}

where $A = 1 - 2L\eta_m$.

Letting $\mathbf{L} =2 L {\eta_m}^2 (2 + A) \Gamma$,  $ \mathbf{M} = 2\eta_m A (\eta_m L - 1) \sum_{i=1}^{n} {(\frac{|D^{i}|}{\sum^{n}_{i=1}|D^{i}|})} (\Phi_i(\bar{\mathbf{w}}_m) - \Phi^*)$

If $ 0 \leq \eta_m \leq \frac{1}{4L}$, then $ 1 \leq A \leq \frac{1}{2}$.

Therefore for term $\mathbf{L} $:
\begin{equation}
	\mathbf{L} = 2 L {\eta_m}^2 (2 + A) \Gamma \leq 6 L {\eta_m}^2 \Gamma
\end{equation}

For term $\mathbf{M} $:
\begin{equation}
	\mathbf{M} = 2\eta_m A (\eta_m L - 1) \sum_{i=1}^{n} {(\frac{|D^{i}|}{\sum^{n}_{i=1}|D^{i}|})} (\Phi_i(\bar{\mathbf{w}}_m) - \Phi^*) < 0
\end{equation}

Back to term $\mathbf{J} $, we have:

\begin{equation}
	\begin{aligned}
		\MoveEqLeft 4L\eta_m^2 \sum_{i=1}^{n} {(\frac{|D^{i}|}{\sum^{n}_{i=1}|D^{i}|})} (\Phi_i(\mathbf{w}^i_m) - \Phi_i^*) - 2\eta_m \sum_{i=1}^{n} {(\frac{|D^{i}|}{\sum^{n}_{i=1}|D^{i}|})} (\Phi_i(\mathbf{w}^i_m) - \Phi_i(\mathbf{w}^*) + \Phi_i^* - \Phi_i^* ) \\
		& \leq A \sum_{i=1}^{n} {(\frac{|D^{i}|}{\sum^{n}_{i=1}|D^{i}|})} \|\mathbf{w}^i_m - \bar{\mathbf{w}}_m\|^2 + 6L\eta_m^2 \Gamma \\
		& \leq \sum_{i=1}^{n} {(\frac{|D^{i}|}{\sum^{n}_{i=1}|D^{i}|})} \|\mathbf{w}^i_m - \bar{\mathbf{w}}_m\|^2 + \frac{3\Gamma}{8L}.
	\end{aligned}
	\label{f}
\end{equation}

Substitute the upper bound of term $\mathbf{J} $ (Eq.\eqref{f}) into Eq.\eqref{e}, we get the bound of $\|\bar{\mathbf{w}}_m - \mathbf{w}^* - \eta_m \bar{\mathbf{g}}_m\|^2$: 

\begin{equation}
	\begin{aligned}
		\MoveEqLeft \|\bar{\mathbf{w}}_m - \mathbf{w}^* - \eta_m \bar{\mathbf{g}}_m\|^2  \\
		&\leq (1-\mu\eta_m) \|\bar{\mathbf{w}}_m - \mathbf{w}^*\|^2 + \sum_{i=1}^{n} {(\frac{|D^{i}|}{\sum^{n}_{i=1}|D^{i}|})} \|\bar{\mathbf{w}}_m - \mathbf{w}^i_m\|_2^2 \\
		&\quad + 4L\eta_m^2 \sum_{i=1}^{n} {(\frac{|D^{i}|}{\sum^{n}_{i=1}|D^{i}|})} (\Phi_i(\mathbf{w}^i_m) - \Phi_i^*) - 2\eta_m \sum_{i=1}^{n} {(\frac{|D^{i}|}{\sum^{n}_{i=1}|D^{i}|})} (\Phi_i(\mathbf{w}^i_m) - \Phi_i(\mathbf{w}^*)) \\
		&\leq (1-\mu\eta_m) \|\bar{\mathbf{w}}_m - \mathbf{w}^*\|^2 + 2 \sum_{i=1}^{n} {(\frac{|D^{i}|}{\sum^{n}_{i=1}|D^{i}|})} \|\bar{\mathbf{w}}_m - \mathbf{w}^i_m\|_2^2 + \frac{3\Gamma}{8L}.
	\end{aligned}
	\label{g}
\end{equation}

\subsubsection{Bounding $\eta_m^2 \|\bar{\mathbf{g}}_m - \mathbf{g}_m\|^2$}

In order to associate with our assumptions, we consider the expectation of $\eta_m^2 \|\bar{\mathbf{g}}_m - \mathbf{g}_m\|^2$:

\begin{equation}
	\begin{aligned}
		\mathbb{E} \|\bar{\mathbf{g}}_m - \mathbf{g}_m\|^2 &= \mathbb{E} \left\| \sum_{i=1}^{n} {(\frac{|D^{i}|}{\sum^{n}_{i=1}|D^{i}|})} (\nabla\Phi_i(\mathbf{w}^i_m, b^i_{m}) - \nabla\Phi_i(\mathbf{w}^i_m)) \right\|^2 \\
		&\leq \mathbb{E} \left( \sum_{i=1}^{n} {(\frac{|D^{i}|}{\sum^{n}_{i=1}|D^{i}|})} \|\nabla\Phi_i(\mathbf{w}^i_m, b^i_{m}) - \nabla\Phi_i(\mathbf{w}^i_m)\| \right)^2 \\
		&= \sum_{i=1}^{n} {(\frac{|D^{i}|}{\sum^{n}_{i=1}|D^{i}|})}^2 \mathbb{E} \|\nabla\Phi_i(\mathbf{w}^i_m, b^i_{m}) - \nabla\Phi_i(\mathbf{w}^i_m)\|^2 \\
		&\leq \sum_{i=1}^{n} {(\frac{|D^{i}|}{\sum^{n}_{i=1}|D^{i}|})}^2 ({\Delta G}_i)^2 \quad \text{(by Assumption 3)}.
	\label{l}
	\end{aligned}
\end{equation}

\subsubsection{Bounding $\mathbb{E} \|\bar{\mathbf{w}}_{m+1} - \mathbf{w}^*\|^2$}

According to Eq.\eqref{i}:
\begin{equation}
	\begin{aligned}
		\|\bar{\mathbf{w}}_{m+1} - \mathbf{w}^*\|^2 
		&= \|\bar{\mathbf{w}}_m  -\mathbf{w}^* - \eta_m \bar{\mathbf{g}}_m \|^2 + \eta_m^2 \|\mathbf{g}_m - \mathbf{g}_m\|^2  + 2\eta_m \langle \bar{\mathbf{w}}_m - \mathbf{w}^* - \eta_m \bar{\mathbf{g}}_m, \bar{\mathbf{g}}_m - \mathbf{g}_m \rangle.
	\end{aligned}
	\label{j}
\end{equation}

We substitute the bound of $\|\bar{\mathbf{w}}_m  -\mathbf{w}^* - \eta_m \bar{\mathbf{g}}_m \|^2$ into Eq.\eqref{j}, we have:
\begin{equation}
	\begin{aligned}
		\|\bar{\mathbf{w}}_{m+1} - \mathbf{w}^*\|^2 
		&\leq (1-\mu\eta_m) \|\bar{\mathbf{w}}_m - \mathbf{w}^*\|^2 + 2 \sum_{i=1}^{n} {(\frac{|D^{i}|}{\sum^{n}_{i=1}|D^{i}|})} \|\bar{\mathbf{w}}_m - \mathbf{w}^i_m\|_2^2 + \frac{3\Gamma}{8L}	\\
		&\quad + \eta_m^2 \|\mathbf{g}_m - \mathbf{g}_m\|^2  + 2\eta_m \langle \bar{\mathbf{w}}_m - \mathbf{w}^* - \eta_m \bar{\mathbf{g}}_m, \bar{\mathbf{g}}_m - \mathbf{g}_m \rangle.
	\end{aligned}
	\label{k}
\end{equation}

Taking the expectation of both sides of Eq.\eqref{k}:
\begin{equation}
	\begin{aligned}
		\mathbb{E} \|\bar{\mathbf{w}}_{m+1} - \mathbf{w}^*\|^2 
		&\leq (1-\mu\eta_m) \mathbb{E} \|\bar{\mathbf{w}}_m - \mathbf{w}^*\|^2 + 2 \sum_{i=1}^{n} {(\frac{|D^{i}|}{\sum^{n}_{i=1}|D^{i}|})} \mathbb{E} \|\bar{\mathbf{w}}_m - \mathbf{w}^i_m\|_2^2 + \frac{3\Gamma}{8L}	\\
		&\quad + \eta_m^2 \mathbb{E} \|\bar{\mathbf{g}}_m - \mathbf{g}_m\|^2  + 2\eta_m \mathbb{E} \langle \bar{\mathbf{w}}_m - \mathbf{w}^* - \eta_m \bar{\mathbf{g}}_m, \bar{\mathbf{g}}_m - \mathbf{g}_m \rangle.
	\end{aligned}
	\label{m}
\end{equation}

According to the definition of $\bar{\mathbf{g}}_m$ and $\mathbf{g}_m$, we have $\bar{\mathbf{g}}_m = \mathbb{E} \mathbf{g}_m$, this implies:
\begin{equation}
	2\eta_m \mathbb{E} \langle \bar{\mathbf{w}}_m - \mathbf{w}^* - \eta_m \bar{\mathbf{g}}_m, \bar{\mathbf{g}}_m - \mathbf{g}_m \rangle = 0
\end{equation}

Substitute Eq.\eqref{l} into Eq.\eqref{m}, we have:
\begin{equation}
	\begin{aligned}
		\MoveEqLeft \mathbb{E} \|\bar{\mathbf{w}}_{m+1} - \mathbf{w}^*\|^2 \\
		&\leq (1-\mu\eta_m) \mathbb{E} \|\bar{\mathbf{w}}_m - \mathbf{w}^*\|^2 + 2 \sum_{i=1}^{n} {(\frac{|D^{i}|}{\sum^{n}_{i=1}|D^{i}|})} \mathbb{E} \|\bar{\mathbf{w}}_m - \mathbf{w}^i_m\|_2^2 + \eta_m^2 \sum_{i=1}^{n} {(\frac{|D^{i}|}{\sum^{n}_{i=1}|D^{i}|})}^2 ({\Delta G}_i)^2 + \frac{3\Gamma}{8L}.
	\end{aligned}
	\label{n}
\end{equation}

Letting $\mathbf{N} = \sum_{i=1}^{n} {(\frac{|D^{i}|}{\sum^{n}_{i=1}|D^{i}|})} \mathbb{E} \|\bar{\mathbf{w}}_m - \mathbf{w}^i_m\|_2^2$

\paragraph{Bounding term $\mathbf{N}$}

Consider the recurrence relationship from $\mathbf{w}^i_m$ to $\bar{\mathbf{w}}_{m_x}$:

\begin{equation}
	\begin{aligned}
		\mathbf{w}^i_m &= \mathbf{w}^i_{m-1} - \eta_{m-1} \nabla\Phi_{i}(\mathbf{w}^i_{m-1}, b_{m-1}) \\
		\mathbf{w}^i_{m-1} &= \mathbf{w}^i_{m-2} - \eta_{m-2} \nabla\Phi_{i}(\mathbf{w}^i_{m-2}, b_{m-2}) \\
		& \vdots \\
		\mathbf{w}^i_{{m_x}+1} &= \bar{\mathbf{w}}_{m_x} - \eta_{m_x} \nabla\Phi_{i}(\mathbf{w}^i_{m_x}, b_{m_x}) \\
		\Rightarrow \bar{\mathbf{w}}_{m_x} - \mathbf{w}^i_m & = \sum_{k=m_x}^{m-1} \eta_{k} \nabla\Phi_{i}(\mathbf{w}^i_{k}, b^i_{k}) 
	\end{aligned}
	\label{o}		
\end{equation}

In $E_{re}$ epochs of the local model regularization of participant $P_i$, we consider the $m_x$-th mini-batch SGD which satisfies $m_x \leq m$ and $m - m_x \leq E_{re} - 1$.

For term $\mathbf{N}$, we have:

\begin{equation}
	\begin{aligned}
		\sum_{i=1}^{n} {(\frac{|D^{i}|}{\sum^{n}_{i=1}|D^{i}|})} \mathbb{E} \|\bar{\mathbf{w}}_m - \bar{\mathbf{w}}_i\|^2 &= \sum_{i=1}^{n} {(\frac{|D^{i}|}{\sum^{n}_{i=1}|D^{i}|})} \mathbb{E} \|\left(\bar{\mathbf{w}}_m - \bar{\mathbf{w}}_{m_x}\right) - \left(\mathbf{w}^i_m - \bar{\mathbf{w}}_{m_x}\right)\|^2 \\
		&= \sum_{i=1}^{n} {(\frac{|D^{i}|}{\sum^{n}_{i=1}|D^{i}|})} \mathbb{E} \|\left(\mathbf{w}^i_m - \bar{\mathbf{w}}_{m_x}\right) - \mathbb{E} \left(\mathbf{w}^i_m - \bar{\mathbf{w}}_{m_x}\right)\|^2 \\
		&\leq \sum_{i=1}^{n} {(\frac{|D^{i}|}{\sum^{n}_{i=1}|D^{i}|})} \mathbb{E} \|\left(\mathbf{w}^i_m - \bar{\mathbf{w}}_{m_x}\right)\|^2 \\
		&\leq \sum_{i=1}^{n} {(\frac{|D^{i}|}{\sum^{n}_{i=1}|D^{i}|})} \mathbb{E} \left\| \sum_{k=m_x}^{m-1} \eta_k \nabla\Phi_i(\mathbf{w}^i_k, b^i_k) \right\|^2 \quad \text{(by Eq.\eqref{o})} \\
		&\leq (m - m_x)^2 \sum_{i=1}^{n} {(\frac{|D^{i}|}{\sum^{n}_{i=1}|D^{i}|})} \mathbb{E} \left\| \sum_{k=m_x}^{m-1} \eta_k \nabla\Phi_i(\mathbf{w}^i_k, b^i_k) \right\|^2 \\
		&\leq (E_{re} - 1)^2 \sum_{i=1}^{n} {(\frac{|D^{i}|}{\sum^{n}_{i=1}|D^{i}|})} {\eta^2_{m_x}} {G^2}\\
		&\leq 4 \eta_m^2 (E_{re} - 1)^2 G^2	\quad \text{(by $\eta_m \leq \eta_{m_x} \leq 2\eta_m$)}.
	\end{aligned}
	\label{p}
\end{equation}

By plugging Ineq.\eqref{p} into Ineq.\eqref{n}, we get:
\begin{equation}
	\begin{aligned}
		\mathbb{E} \|\bar{\mathbf{w}}_{m+1} - \mathbf{w}^*\|^2 
		&\leq (1-\mu\eta_m) \mathbb{E} \|\bar{\mathbf{w}}_m - \mathbf{w}^*\|^2 + 8 \eta_m^2 (E_{re} - 1)^2 G^2 + \eta_m^2 \sum_{i=1}^{n} {(\frac{|D^{i}|}{\sum^{n}_{i=1}|D^{i}|})}^2 ({\Delta G}_i)^2 + \frac{3\Gamma}{8L}.
	\end{aligned}
\end{equation}

Letting $\delta_{m} = \mathbb{E} \|\bar{\mathbf{w}}_m - \mathbf{w}^*\|^2$, and thus $\delta_{m+1} = \mathbb{E} \|\bar{\mathbf{w}}_{m+1} - \mathbf{w}^*\|^2$, we can get the following recurrence relation:
\begin{equation}
	\begin{aligned}
		\delta_{m+1}
		&\leq (1-\mu\eta_m) \delta_{m} + 8 \eta_m^2 (E_{re} - 1)^2 G^2 + \eta_m^2 \sum_{i=1}^{n} {(\frac{|D^{i}|}{\sum^{n}_{i=1}|D^{i}|})}^2 ({\Delta G}_i)^2 + \frac{3\Gamma}{8L}.
	\end{aligned}
	\label{q}
\end{equation}

$\delta_{m}$ stands for the expectation of the difference between global model in the $m$-th mini-batch SGD $\bar{\mathbf{w}}_m$ and the optimal global model $w^*$. 

In order to guarantee the convergence of array $\{\delta_m\}$, the following conditions should be satisfied:
\begin{enumerate}
	\item The learning rate $\eta_m$ decays with the number of mini-batch SGD.
	\item $0 < (1 - \mu\eta_m) < 1$
	\item $(E_{re} - 1)^2 G^2$, $({\Delta G}_i)^2$ and $\frac{3\Gamma}{8L}$ are bounded.
\end{enumerate}

Combining our assumptions, these conditions are all satisfied. Therefore, $\delta_m$ convergences to 0 as $m$ increases, which implies that the global model $\bar{\mathbf{w}}_m$ will convergence to the optimal global model $w^*$.
	
\end{appendices}

\end{document}